\documentclass[journal,onecolumn]{IEEEtran}
\IEEEoverridecommandlockouts
\usepackage{cite}
\usepackage{amsmath,amssymb,amsfonts}
\usepackage{graphicx}
\usepackage{textcomp}
\usepackage{hyperref}
\usepackage{xcolor}
\usepackage{url}
\def\BibTeX{{\rm B\kern-.05em{\sc i\kern-.025em b}\kern-.08em
    T\kern-.1667em\lower.7ex\hbox{E}\kern-.125emX}}

\usepackage{multirow}
\usepackage{algorithm,algpseudocode}
\usepackage{comment}
\usepackage{amsthm}

\newcommand{\norm}[1]{\lVert#1\rVert}               

\DeclareMathOperator*{\argmin}{arg\,min}

\makeatletter
\def\BState{\State\hskip-\ALG@thistlm}
\makeatother

\begin{document}

\title{Byzantine-Robust Aggregation for Securing Decentralized Federated Learning\\
}

\author{
    \IEEEauthorblockN{Diego Cajaraville-Aboy\textsuperscript{*}, Ana Fernández-Vilas, Rebeca P. Díaz-Redondo, Manuel Fernández-Veiga}\\
    \IEEEauthorblockA{atlanTTic Research Center -- ICLAB -- Universidade de Vigo, Vigo, 36310, Spain\\
    \{dcajaraville,avilas,rebeca,mveiga\}@det.uvigo.es}\\
    \textsuperscript{*}{Corresponding author: dcajaraville@det.uvigo.es}%
    \thanks{This version of the article has been accepted for publication in IEEE Access after peer review, but is not the Version of Record and may not reflect final copy-editing, formatting, pagination, or corrections. The Version of Record is available online at: \url{https://doi.org/10.1109/ACCESS.2025.3629864}. © 2025 The Authors. This manuscript version is licensed under CC BY 4.0.}%
}

\maketitle

\begin{abstract}
Federated Learning (FL) emerges as a distributed machine learning approach that addresses privacy concerns by training AI models locally on devices. Decentralized Federated Learning (DFL) extends the FL paradigm by eliminating the central server, thereby enhancing scalability and robustness through the avoidance of a single point of failure. However, DFL faces significant challenges in optimizing security, as most Byzantine-robust algorithms proposed in the literature are designed for centralized scenarios. In this paper, we present a novel Byzantine-robust aggregation algorithm to enhance the security of Decentralized Federated Learning environments, coined \textsf{WFAgg}. This proposal handles adverse conditions and strengthens the robustness of dynamic decentralized topologies at the same time by employing multiple filters to identify and mitigate Byzantine attacks. Experimental results demonstrate the effectiveness of the proposed algorithm in maintaining model accuracy and convergence in the presence of various Byzantine attack scenarios, outperforming state-of-the-art centralized Byzantine-robust aggregation schemes (such as \textsf{Multi-Krum} or \textsf{Clustering}). These algorithms are evaluated on an IID image classification problem in both centralized and decentralized scenarios.
\end{abstract}


\section{Introduction}
\label{sewc:introduction}

The integration of Machine Learning (ML) into modern technology has significantly transformed various sectors, particularly with the proliferation of Internet of Things (IoT) devices~\cite{Mbock20}. This growth has led to an exponential increase in the generation and distribution of big data across devices, highlighting both new opportunities and challenges for advanced data processing techniques~\cite{Suganyadevi22}. The classical approach to handling this amount of data has several disadvantages in terms of security, privacy, and scalability, as all this data (and the trained algorithm) is stored and processed in a large centralized datacenter~\cite{Shokri15,Liu20}. In response to these challenges, Federated Learning (FL) has been proposed as an alternative paradigm that allows machine learning algorithms to work without the centralization of data~\cite{McMahan17}. FL leverages the computational capabilities of end devices, enabling them to train their own ML models with their local data. A central server then merges all the locally trained models from the end devices to compute a global model~\cite{Bonawitz17}. This method enhances privacy by design, as it does not expose all data at a single location, while utilizing the processing capabilities of the devices. However, traditional federated learning still relies on a central server to manage the learning algorithm process, which can be a target for privacy and security attacks by malicious attackers~\cite{Truex19}, among other ongoing challenges.

The concept of Decentralized Federated Learning (DFL) extends FL principles by eliminating the central server, thereby mitigating the risks associated with a single point of failure and enhancing privacy by avoiding a centralized coordination of the learning process~\cite{Yuan24}. This decentralized approach promotes greater resilience, allowing the learning process to continue even if some nodes fail or go offline, thus improving the robustness and fault tolerance of the overall network~\cite{Li22}.

However, while the decentralization of these learning environments offers benefits such as improved scalability, it also introduces new challenges, including increased vulnerability to security threats. The absence of a central server complicates the implementation of uniform security-preserving algorithms and makes the system more susceptible to attacks such as Byzantine failures~\cite{Lyu20, Fang21}. Notably, Byzantine failures occur when a group of Byzantine clients attempts to disrupt the proper learning process, e.g., by altering their updates to degrade other local model's performance (known as Byzantine attacks~\cite{Marano09}). Additionally, the heterogeneity among nodes, with varying computational capabilities or data quality, which can lead to inconsistencies and reduce the effectiveness of the learned model~\cite{Pandi23}.

Despite the potential of the DFL paradigm, research on securing such environments is still limited, with most existing work focusing on centralized FL approaches~\cite{Mothukuri21}. Many of these approaches utilize Byzantine-robust aggregation algorithms designed to mitigate/filter malicious attacks from adversarial nodes in the central server~\cite{Blanchard17, Pillutla22}. This raises the question: are these algorithms proposed for centralized scenarios equally efficient in decentralized environments?

In light of these challenges, we introduce an innovative Byzantine-robust aggregation algorithm designed to ensure the proper performance of Decentralized Federated Learning environments, which we coined \textsf{WFAgg}. To the best of our knowledge, this represents a novel approach in the literature, as existing works have not explicitly addressed Byzantine robustness under decentralized scenarios considering both spatial and time-varying statistics. The proposed algorithm leverages multiple techniques to filter and mitigate Byzantine attacks during the aggregation step in DFL, adapting effectively to these conditions. In summary, the contributions of this paper are as follows:

\begin{itemize}
    \item We design and develop a novel Byzantine-robust aggregation algorithm for DFL environments. It ensures robustness against malicious attacks by using different lightweight filtering techniques, which is particularly suited for adversarial decentralized IoT environments.
    
    \item We systematically evaluate different state-of-the-art (SOTA) centralized Byzantine-robust aggregation algorithms in decentralized settings and compare these results to those obtained in the centralized ones.
    
    \item We assess the robustness of the proposed algorithm through multiple experiments on an image classification problem under different Byzantine attacks. This algorithm obtains better accuracy and model consistency results compared with the centralized ones under the evaluated tests.

    \item We implemented a DFL simulator in the Python language to conduct the aforementioned experiments.
\end{itemize}

The remainder of this paper is structured as follows. Section~\ref{sec:review} includes a detailed explanation of concepts such as Federated Learning and Decentralized Federated Learning, focusing on the current state of the art of the latter. Section~\ref{sec:security} provides a study of security in FL environments, along with the most well-known techniques for addressing the associated challenges. Section~\ref{sec:methodology} presents the proposed Byzantine-robust aggregation algorithm solution in this paper, providing context for the scenario addressed and details about employed techniques and mechanisms. Section~\ref{sec:implementation} details the simulation settings, including configuration parameters and specifications. Then, Section~\ref{sec:results} discusses the obtained results regarding the system's robustness and convergence (both in the proposed algorithm and in the SOTA algorithms for centralized schemes). Finally, Section~\ref{sec:conclusions} offers the final conclusions of this paper, highlighting the contributions to the field of DFL research and suggesting possible future lines of work.

\section{Background}
\label{sec:review}

Originally introduced by Google in 2016~\cite{McMahan17}, the classical version of federated learning involves each client maintaining a local model trained with its own training dataset, while a central aggregator (or parameter server) maintains a global model that is updated through the local models of the clients. In depth, centralized FL proceeds as follows~\cite{Li21} (in a specific round): (i) the server sends the current global model parameters to all client devices, (ii) each client updates its local model by performing one or several steps of an optimization algorithm (in the literature, the most common is Stochastic Gradient Descent) using the current global model parameters and their local training datasets, then, (iii) devices send their updated local models back to the server where (iv) the server aggregates these local models from the clients using a specific aggregation rule to compute a new global model for the next round. The most recognized aggregation rule in FL field is Federated Average \textsf{FedAvg}. This process is carried out in multiple communication rounds, an iterative process that continues until the global model reaches the desired level of accuracy. 

Of particular interest in this work is the novel paradigm of Decentralized Federated Learning which, despite not being the predominant approach in this field, introduces certain features that make it a promising alternative. In DFL, various nodes/clients collaborate and train their local models collectively without the necessity of a central server~\cite{Kairouz21,He18}, distributing the aggregation process among the nodes of the learning network. This eliminates the need to share a global model with all clients~\cite{Geiping20}.

One of the key advantages of DFL is promoting scalable and resilience environments, allowing nodes to join or disconnect from the learning network based on the power resources of the devices. Unlike centralized approaches, which can become bottlenecks or single points of failure, DFL distributes the workload over the entire network, improving fault tolerance and making the system more robust against attacks~\cite{Beltran23}. Another significant advantage of DFL over centralized approaches is the reduction in communication resources and the decrease in high bandwidth usage due to the elimination of intermediate model/gradient transmissions between clients and the server~\cite{Yuan24}. This is particularly relevant in environments with a large number of interconnected devices.

Despite the significant potential of DFL, research on this topic remains sparse, indicating it is an emerging field. This scenario must confront new challenges absent in centralized approaches due to the decentralized learning process and decision-making across numerous devices. For example, designing communication protocols that define how clients exchange their models in order to prevent excessive communication overheads. Gossip Learning is a notable decentralized communication protocol for asynchronous updates between nodes, leveraging peer-to-peer (P2P) schemes to achieve scalability and potential privacy preservation~\cite{Ormandi13, Prabhakar22}. Although effective, gossip learning has a slower aggregation evolution due to its fully distributed nature~\cite{Yuan24}. Hybrid protocols combining gossip and broadcast techniques offer greater adaptability to various decentralized scenarios, facilitating model sharing only with neighboring nodes~\cite{Abdelghany22}. Other works~\cite{Beltran23, Nguyen21} propose blockchain-based FL environments that provide trustworthy schemes by guaranteeing the immutability of exchanged models through a peer-to-peer consensus mechanism. 

Moreover, DFL approaches lead to high inconsistency among local models since there is no global model. Some proposals, such as increasing gossip steps in local communications, achieve better consensus~\cite{Shi23}. Another approach involves a consensus-based DFL algorithm inspired by discrete-time weighted average consensus frameworks~\cite{Giuseppi22}. Additionally, techniques like optimizing the consensus matrix to enhance the convergence rate and architectures focusing on continuous consensus over aggregated models are explored~\cite{Du23,Schmid20}.

\section{Related Work}
\label{sec:security}

This section provides an overview of related work in the state of the art, highlighting key areas of security in Federated Learning. It is divided into three subsections that review security vulnerabilities, existing Byzantine-robust aggregation schemes proposed for centralized scenarios, and related works on various Byzantine-robust aggregation scheme proposals.

\subsection{Security Vulnerabilities}

The FL process is vulnerable to attacks on collaborative training, which can negatively impact model performance and the overall robustness of the system. Model performance can be compromised through specific targeted attacks (including backdoor attacks that disrupt the model's ability to accurately classify a particular category) or through general untargeted attacks (such as Byzantine attacks, which focus on obstructing the collaborative training process by providing false updates) \cite{Neto23}. There are two main different ways of carrying out untargeted attacks: by poisoning the data with which the model is trained (data poisoning) or by manipulating the local model before sending it to the server (model poisoning). As previous research~\cite{Wang20} has shown that model poisoning attacks have a greater impact compared to data poisoning attacks, this work will concentrate on the former.

One of the most well-known techniques to counter data/model poisoning attacks performed by malicious nodes, which can be considered Byzantine attacks, are Byzantine-robust techniques. To mitigate or eliminate these Byzantine faults that threaten the proper functioning of FL, various Byzantine-robust algorithms have 
emerged~\cite{Shi22}. These include redundancy-based and trust-based schemes that seek to mitigate Byzantine attacks either by assigning each client redundant updates or by assuming the presence of reliable nodes for filtering attacks, respectively~\cite{Rajput20,Park21}. Of significant importance are robust aggregation schemes that aim to optimize the aggregation process itself to mitigate or filter Byzantine attacks~\cite{Blanchard17,Li21,Yin18,Sattler20,Li24}.

\subsection{Main Centralized Byzantine-Robust Algorithms}

Due to the importance they will have throughout this document, we are going to describe some of the most renowned Byzantine-robust aggregation schemes for CFL in the state of the art. Suppose a server receives a subset of models $\{\theta_k\}_{k=1}^{K}$ from its $K$ clients.

\begin{itemize}
    \item \textsf{Mean} \cite{McMahan17}: This is the most basic aggregation rule and often serves as a benchmark for comparisons. This algorithm aggregates client models by calculating the coordinate-wise arithmetic mean of the given set of models. It simplifies the \textsf{FedAvg} algorithm which uses a weighted average based on the number of samples trained by each client.

    \item \textsf{Median} \cite{Yin18}: This algorithm is based on a coordinate-wise aggregation rule to compute the global model. To achieve this, the server calculates each $i$-th parameter of the model by sorting the $K$ values from the received models and computing their median value. 

    \item \textsf{Trimmed-Mean} \cite{Yin18}: This is a variant of the \textsf{Mean} aggregation algorithm and another coordinate-wise aggregation rule. For a given trim rate $\beta \in \left ( 0, 1/2 \right )$, the server sorts the $K$ received values of each $i$-th parameter in the model and removes the smallest and largest $\beta K$ values. So, the $i$-th parameter of the global model is computed as the mean of the remaining $(1-2\beta)K$ values.

    \item \textsf{Krum} and \textsf{Multi-Krum} \cite{Blanchard17}: This algorithm filters the models received from clients using Euclidean distances. For each client, the server assigns a score by calculating the sum of Euclidean distances from the client's model $\theta_k$ to the $K-M-2$ closest neighbours models, where $M$ is the number of malicious models. The model with the lowest score is selected as the global model. The \textsf{Multi-Krum} variant selects the $m$ models with the lowest scores and computes the global model as the mean of these $m$ selected models.
    
    \item \textsf{Clustering} \cite{Li21,Sattler20}: The aim of this algorithm is to separate the models received from clients into two clusters and aggregate as the global model those belonging to the larger cluster. This is achieved using an agglomerative clustering algorithm with average linkage, employing pairwise cosine similarities between models as a distance metric.
    
\end{itemize}

The majority of existing Byzantine-robust aggregation schemes (including those described previously) can be categorized based on the techniques employed by them~\cite{Shi22}:
\begin{itemize}
\item \textbf{Statistics:} These schemes utilize the statistical characteristics of updates, such as the median, to block abnormal updates and achieve robust aggregation. They are suitable only when the number of malicious users is less than half of the total users.
\item \textbf{Distance:} These schemes aim to detect and discard bad or malicious updates by comparing the distances between updates. An update that is significantly different from others is considered malicious. These are suitable only to resist attacks that produce noticeably abnormal updates.
\item \textbf{Performance:} In these schemes, each update is evaluated over a clean dataset provided by the server, and any update that performs poorly is given low weight or removed directly. These schemes are more reliable than other solutions but rely on a clean dataset for evaluations, which is not always possible.
\end{itemize}

\subsection{FL Defense Schemes} 

Following the previous subsection, not all proposals in the state of the art fit into the aforementioned classification. This is due to the development of innovative techniques that offer different approaches compared to the more well-known aggregation algorithms.

\textsc{SignGuard}~\cite{Xu23} is a novel approach which enhances FL system robustness against model poisoning attacks by utilizing the element-wise sign of gradient vectors. It proposes processing the received gradients to generate relevant magnitude, sign, and similarity statistics, which are then used collaboratively by multiple filters to eliminate malicious gradients before the final aggregation. The key idea is that the sign distribution of sign-gradient can provide valuable information in detecting advanced model poisoning attacks.

\textsc{FLTrust}~\cite{Cao22} proposes an alternative method by implementing a trust-based approach. Unlike the previous proposal, this method uses an initial trust model, trained on a small maintained dataset, to evaluate and weight the contributions of clients in the global model aggregation process. Thus, client updates that align closely with the server model's direction are deemed trustworthy, while others are adjusted or discarded based on their deviation.

There are also other proposals that do not only focus on the geometric properties of models for mitigating Byzantine attacks. For example,~\cite{Li21} proposes a method to mitigate the impact of Byzantine clients in federated learning through a spatial-temporal analysis approach. Apart from leveraging clustering-based methods to detect and exclude incorrect updates based on their geometric properties in the parameter space, the temporal analysis provides a more effective defense mechanism, which is crucial in environments where attack strategies may evolve during the course of model training. In this case, time-varying behaviors are addressed through an adjustment of the learning rate parameter.

Another example is \textsc{Siren}~\cite{Guo21} which proposes a FL system to improve robustness through proactive alarming. Clients alert abnormal behaviors in real-time using a continuous monitoring scheme based on the evolution of the global model's accuracy over the rounds. When a potential threat is identified, the clients issue an alarm, and the central server adjusts the model aggregation to mitigate the impact of malicious contributions.

\section{Methodology}
\label{sec:methodology}
  
Despite the existence of multiple Byzantine-robust proposals in the state of the art, the majority have been oriented towards centralized FL scenarios. To our knowledge, there is no proposed Byzantine-robust aggregation algorithm specifically designed to enhance security in DFL environments. In this section, we elaborate on the system configuration employed in DFL and detail all technical specifications of the proposed algorithm \textsf{WFAgg} designed to ensure Byzantine-robustness within this framework. Additionally, the corresponding filtering and aggregation algorithms that are integral components of the \textsf{WFAgg} algorithm will be described in the subsequent subsections.

\subsection{System Setting}

\begin{figure*}[t]
    \centering
    \includegraphics[width=1.0\textwidth]{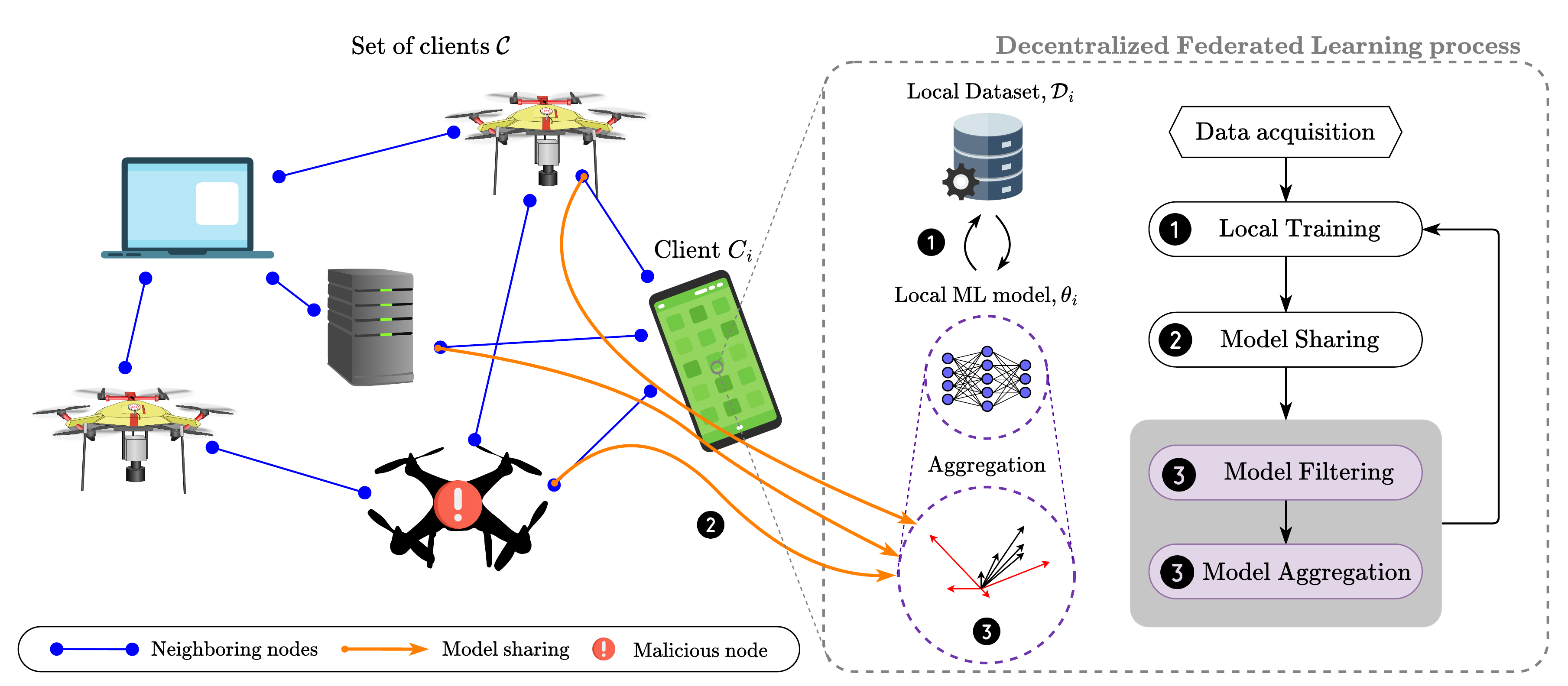}
    \caption{Decentralized Federated Learning framework architecture.}
    \label{fig:scenario}
\end{figure*}

We consider a cross-device scenario where multiple IoT devices (or servers) carry out a distributed collaborative learning of a machine learning model through DFL. Figure~\ref{fig:scenario} illustrates the proposed scenario for the DFL system. This figure showcases a set of clients/devices $\mathcal{C} =\{C_i\}, \ i = 1, 2, \dots, N$ involved in the collaborative training of a ML model. Each client possesses a local dataset $\mathcal{D}_i, i \in [N]$, with $n_i$ data samples, which is used for training their own local model, $\theta_i$. The objective of FL can be formulated as a non-convex minimization problem:
\begin{align}
    \min_{\theta \in \mathbb{R}^d} \mathcal{L}(\theta, \mathcal{D})  = \min_{\theta \in \mathbb{R}^d} \biggl \{ \sum_{i \in [N]} \dfrac{|\mathcal{D}_i|}{|\mathcal{D}|} \mathcal{L}(\theta, \mathcal{D}_i) \biggr \},
\end{align}
where $\mathcal{L}(\theta, \mathcal{D})$ is the empirical loss function of the model $\theta$ over the overall dataset $\mathcal{D} = \bigcup_{i \in [N]} \mathcal{D}_i$, and $|\mathcal{D}| = \sum_{i \in [N]} |\mathcal{D}_i|$ is the total number of samples across all clients.

The topology of the DFL scenario is modeled as an undirected graph $G=(\mathcal{C},\mathcal{E})$ where $\mathcal{C}$ represents the devices participating in the learning process as vertices, which are interconnected by the set of links $\mathcal{E} \subseteq \mathcal{C} \times \mathcal{C}$. Although the topology is dynamic across rounds, the problem is simplified by assuming a static topology where each node consistently communicates with the same set of nodes.

The majority of the client set $\mathcal{C}$ consists of benign clients, i.e., they follow the established protocol and send honest models to the corresponding nodes, while a small fraction are Byzantine clients, i.e., they act maliciously and send arbitrary models with the intent of disrupting the proper training of the model. Nevertheless, this fraction of malicious nodes is relative since each node communicates with a different subset of nodes.

After the training process on its own dataset, each node shares its local model with other nodes in the network. The communication protocol proposed for this framework is based on a Gossip-inspired protocol. Instead of sharing its model with only one random node in the topology, each client forwards the trained model to all of its neighbors in the communication network. The set of neighboring clients of client $C_i$ in topology graph $G$ is defined as $\mathcal{N}_i(G) = \left \{ C_j \in \mathcal{C}, C_j \neq C_i \ : \ \{ C_i, C_j\} \in \mathcal{E} \right \}$.

Upon receiving the trained models from neighboring nodes (assuming receipt of all models), each node applies a Byzantine-robust aggregation algorithm in order to update its local model. This process is repeated iteratively, beginning again with the local model training in the next round.

\subsection{Proposed Byzantine-Robust Aggregation Scheme}

\begin{figure*}[t]
    \centering
    \includegraphics[width=0.9\textwidth]{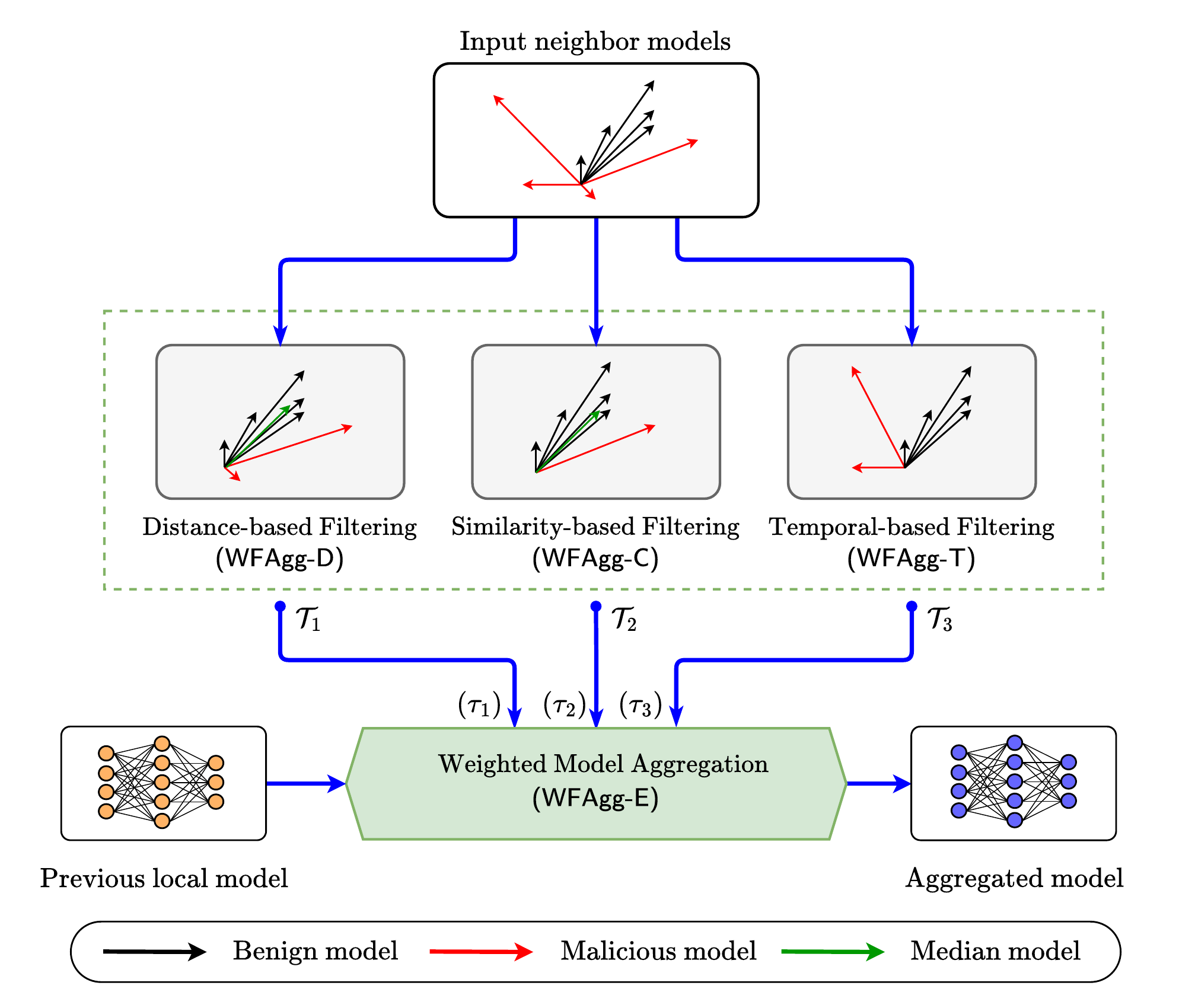}
    \caption{Workflow of Byzantine-robust aggregation algorithm \textsf{WFAgg}.}
    \label{fig:byzantine-robust-algorithm}
\end{figure*}

This proposal suggests, similarly to other approaches, performing a preliminary filtering of the received models before aggregating them. The \textsf{WFAgg} algorithm defines both procedures, and this can be seen in Figure~\ref{fig:byzantine-robust-algorithm} that summarizes the workflow of the Byzantine-robust aggregation algorithm.

In the \textsf{WFAgg} algorithm, the set of received models $\mathcal{T}$ by a client (from its neighboring nodes) is passed through multiple filters, each of them with a different purpose. The aim is to perform a distinct type of criteria in each filter using various well-known techniques in this field for mitigating or eliminating malicious models. This approach seeks to compensate for the inadequacies of some techniques in detecting specific attacks by leveraging other techniques. This fact is especially relevant in adversarial decentralized environments, as the number of received models can be low and the fraction of malicious nodes can be high (depending on the network topology).

Specifically, the algorithm \textsf{WFAgg} comprises three multipurpose filters, including a distance-based filter, a similarity-based filter, and a temporal-based filter. As reviewed in Section~\ref{sec:review}, several Byzantine-robust aggregation algorithms align with the described characteristics and could be utilized (e.g., \textsf{Multi-Krum} or \textsf{Clustering}). However, others such as those based on centrality statistics do not perform a filtering of the models \textit{per se} and therefore would not be suitable (e.g., \textsf{Median}). Despite this, new algorithms for each of these filters, \textsf{WFAgg-D}, \textsf{WFAgg-C} and \textsf{WFAgg-T}, will be presented and detailed in the following subsections.

\begin{algorithm}[t]
\caption{$\mathsf{WFAgg}$: Byzantine-Robust Aggregation Algorithm}
\label{alg:WFAgg}
\begin{algorithmic}[1]
\BState \textbf{Input:} Set of $K$ received updates $\mathcal{T} = \{\theta_j\}_{j \in [K]}$, the number of estimated malicious nodes $f$, length of the time window $W$, $T_\mathsf{th}$ rounds of the transient period and $\alpha$ smoothing factor
\BState \textbf{Output:} Set of benign updates $(\mathcal{T}_1, \mathcal{T}_2, \mathcal{T}_3)$ and aggregated model $\overline{\theta}_i$
\BState \textbf{Initial:} $\mathcal{T}_1 = \mathcal{T}_2 = \mathcal{T}_3 = \emptyset$, $w_{ij} = 0$\\

\State  $\mathcal{T}_1 = \textsf{WFAgg-D}(\mathcal{T}, f)$ 

\State  $\mathcal{T}_2 = \textsf{WFAgg-C}(\mathcal{T}, f)$

\State  $\mathcal{T}_3 = \textsf{WFAgg-T}(\mathcal{T}, W, T_\mathsf{th})$ \\

\For{each $j \in [K]$} 
    \If{$\theta_j \in \mathcal{T}_1$}
        \State $w_{ij} \gets w_{ij} + \tau_1$
    \EndIf
    \If{$\theta_j \in \mathcal{T}_2$}
        \State $w_{ij} \gets w_{ij} + \tau_2$
    \EndIf
    \If{$\theta_j \in \mathcal{T}_3$}
        \State $w_{ij} \gets w_{ij} + \tau_3$
    \EndIf
    \If{$w_{ij} < \min_{\{k,\ell\}, k \neq \ell} (\tau_k + \tau_\ell)$}
        \State $w_{ij} \gets 0$
    \EndIf
\EndFor\\

\State $\overline{\theta}_i = \textsf{WFAgg-E}(\mathcal{T}, \alpha, \{w_{ij}\}_{j\in[K]})$ 

\end{algorithmic}
\end{algorithm}

Once each of these filters has selected a subset of the received models identified as benign ($\mathcal{T}_1, \mathcal{T}_2, \mathcal{T}_3$, respectively for each filter), the model aggregation process will be performed through a weighting of the different filter results. Specifically, each model is scored based on which filters, if any, have identified it as benign. The weighting assigned by each filter is predefined based on the relative importance attributed to that technique in filtering malicious models ($\tau_1, \tau_2, \tau_3$, respectively for each filter, so $\sum_i \tau_i = 1$).

Considering that the algorithm may be attacked by targeting specific characteristics of some filters, it is understandable that a model must be accepted by two or more filters to be considered benign. This design decision is also influenced by the temporal-based filter, as it, by considering only temporal statistics, might identify a constant attack by a malicious node as benign.

Several examples are proposed for better understanding (suppose client $C_i$ is performing the algorithm). If a model $\theta_j$, from one of its neighboring nodes, has been identified as benign by all three filters, it receives the highest weighting in the aggregation process and will be more relevant ($w_{ij} = \tau_1+\tau_2+\tau_3$). Now suppose that the same model is identified as benign by the distance-based and similarity-based filters, so it will receive a lower weighting than in the previous case, resulting in less relevance in the aggregated model ($w_{ij} = \tau_1+\tau_2$). Conversely, if the model is identified as benign only by the similarity-based filter will not be considered in the aggregation process, i.e., it is assigned a zero weight ($w_{ij} = 0$).

The output of the filtering process consists of the triplet $(\mathcal{T}_1, \mathcal{T}_2, \mathcal{T}_3)$, representing the set of models considered benign in this round by each of the filters. Using the weights assigned to each model and the client's own model, a coordinate-wise weighted aggregation (such as \textsf{FedAvg}) is performed, with the difference that this algorithm assigns a differential value to the local model. This proposed aggregation algorithm is called \textsf{WFAgg-E} and will be presented and described in the following subsections.

In summary, Algorithm~\ref{alg:WFAgg} describes the main characteristics of the algorithm functionality. Since the filters operate sequentially and share intermediate model statistics, the overall cost increases only linearly with the number of filters. Although \textsf{WFAgg} incorporates multiple filters, its computational overhead will remain moderate if each filtering step (distance-, similarity-, and temporal-based) has a comparable order of complexity to classical single-filter methods such as \textsf{Krum}. This complexity will be studied in following subsections.

\subsection{WFAgg-D: Distance-based Model Filtering}

The \textsf{WFAgg-D} algorithm (which is described in detail in Algorithm \ref{alg:WFAgg-D}) is proposed as a Byzantine-robust aggregation algorithm aimed at detecting and mitigating models that geometrically differ from the rest. Specifically, it seeks to discard models that are at a greater Euclidean distance from a reference model. This reference model is characterized by the central statistical characteristics of the received models, specifically the information provided by the median model of the set of models.

First, the median (using the \textsf{Median} algorithm explained in Section~\ref{sec:review}) of the set of models received from neighboring nodes is calculated. Then, all Euclidean distances between each of the neighbor models and the reference model are computed. From these metrics, the $K-f-1$ models with the smallest Euclidean distance to the reference model, $\mathcal{T}_1$, are selected, where $K$ is the number of received models and $f$ is the number of malicious nodes.

The computational complexity of the algorithm is $\mathcal{O}(d K \log K)$ for the median calculation since it involves sorting the $K$ values of the received models, $\mathcal{O}(K \log K)$, for each of the $d$ components of the model. The complexities of calculating Euclidean distances and selecting the best results correspond to $\mathcal{O}(d K)$ and $\mathcal{O}(K)$ (on average, using a selection algorithm like \texttt{Quickselect}), respectively, which are terms dominated by the median calculation (for large values of $K, d$).

\begin{algorithm}[tbp]
\caption{\textsf{WFAgg-D}: Distance-based filtering algorithm}
\label{alg:WFAgg-D}
\begin{algorithmic}[1]
\BState \textbf{Input:} Set of $K$ received updates $\{\theta_j\}_{j \in [K]}$, the number of estimated malicious nodes $f$
\BState \textbf{Output:} Set of benign updates $\mathcal{T}_1$
\BState \textbf{Initial:} $\mathcal{T}_1 = \emptyset$\\

\BState \textbf{Filter 1:} Distance-based filtering
\State $\theta_\textsf{Med} = \textsf{Median}(\{\theta_j \ : \ j \in [K]\})$
\For{each $j \in [K]$, }
    \State $d_j =  \| \theta_j - \theta_\textsf{Med} \|^2 $  \Comment{Euclidean distance}
\EndFor\\
\State  $\mathcal{T}_1 = \argmin_{\mathcal{J} \subset [K], |\mathcal{J}| = K-f-1}{ \sum_{j \in \mathcal{J}} d_j } $ 
\end{algorithmic}
\end{algorithm}

\subsection{WFAgg-C: Similarity-based Model Filtering}

The \textsf{WFAgg-C} algorithm (which is described in detail in Algorithm \ref{alg:WFAgg-C}) is proposed as another Byzantine-robust aggregation algorithm with the same objective as the \textsf{WFAgg-D} algorithm but a different criteria. Unlike \textsf{WFAgg-D}, which explores the Euclidean distances between models, \textsf{WFAgg-C} seeks to discard models that represent a significant change in direction relative to a reference model. Similarly, this reference model is characterized by the median of the set of received models. To quantify these directional changes the cosine distance is employed, a metric that measures the similarity between two vectors in a multidimensional space, defined as
\begin{align}
\alpha_{\mathbf{x},\mathbf{y}} := 1 - \cos(\angle\{\mathbf{x},\mathbf{y}\}) = 1 - \dfrac{\langle \mathbf{x},\mathbf{y} \rangle}{\norm{\mathbf{x}} \cdot \norm{\mathbf{y}}},
\end{align}
for two vectors $\mathbf{x}, \mathbf{y} \in \mathbb{R}^d$. Remark that $\alpha_{\mathbf{x},\mathbf{y}} \in [0,2]$.

First, the median (using the \textsf{Median} algorithm explained in Section~\ref{sec:review}) of the set of models received from neighboring nodes is calculated. Additionally, the median magnitude is calculated to perform magnitude clipping on the models (the magnitude is not a relevant parameter in this algorithm). Then, all cosine distances between each of the neighbor models and the reference model are computed. From these metrics, the $K-f-1$ models with the smallest cosine distance to the reference model, $\mathcal{T}_2$, are selected, where $K$ is the number of received models and $f$ is the number of malicious nodes.
The computational complexity of the algorithm can be denoted as $\mathcal{O}(d K \log K)$. This matches the complexity of the previous algorithm since the calculation of the median model remains the dominant term (the computation of the inner product is linear with $d$).

\begin{algorithm}[ht]
\caption{\textsf{WFAgg-C}: Similarity-based filtering algorithm}
\label{alg:WFAgg-C}
\begin{algorithmic}[1]
\BState \textbf{Input:} Set of $K$ received updates $\{\theta_j\}_{j \in [K]}$, the number of estimated malicious nodes $f$
\BState \textbf{Output:} Set of benign updates $\mathcal{T}_2$
\BState \textbf{Initial:} $\mathcal{T}_2 = \emptyset$\\

\BState \textbf{Filter 2:} Similarity-based filtering
\State $\theta_\textsf{Med} = \textsf{Median}(\{\theta_j \ : \ j \in [K]\})$
\State $\tau_\textsf{Med} = \textsf{Median}(\{ \| \theta_j \| \ : \ j \in [K]\})$\\
\For{each $j \in [K]$ }
    \State $\theta_j' = \theta_j \cdot \min \left ( 1, \dfrac{\tau_\textsf{Med}}{\| \theta_j \|}  \right )$ \Comment{Norm clipping}
    \State $\alpha_j =  1 - \dfrac{\langle \theta'_j, \theta_\textsf{Med}  \rangle}{\| \theta'_j \| \cdot \| \theta_\textsf{Med} \|} $  \Comment{Cosine distance} 
\EndFor\\
\State  $\mathcal{T}_2 = \argmin_{\mathcal{J} \subset [K], |\mathcal{J}| = K-f-1}{ \sum_{j \in \mathcal{J}} \alpha_j } $ 
\end{algorithmic}
\end{algorithm}

\subsection{WFAgg-T: Temporal-based Model Filtering}

The \textsf{WFAgg-T} algorithm (which is described in detail in Algorithm \ref{alg:WFAgg-T}) is proposed as a model filtering algorithm aimed at detecting nodes that make abrupt behavioral changes in models between rounds. It seeks to identify if a node has conducted a Byzantine attack in a particular round or if the nature of such an attack is semi-random, causing sudden model changes between rounds. Therefore, this algorithm cannot be considered for a Byzantine-robust aggregation scheme because if a malicious node perpetrates a well-designed attack with little variability between rounds, will not be detected.

The comparison of a node's current model with its models sent in previous rounds (which should be non consecutive) is done using geometric metrics, specifically, Euclidean distance and cosine similarity. When analyzing the temporal changes between rounds for a node, a time window of the last rounds is considered to extract the necessary metrics. Each node only needs to store the history of the distance metrics and only the last model sent by each neighboring node. Both the mean and the variability within this time window are studied to compare with the latest model update.

\begin{algorithm}[ht]
\caption{\textsf{WFAgg-T}: Temporal-based filtering algorithm}
\label{alg:WFAgg-T}
\begin{algorithmic}[1]
\BState \textbf{Input:} Set of $K$ received updates $\{\theta_j\}_{j \in [K]}$, length of the time window $W$, current communication round $t$ and $T_\mathsf{th}$ rounds of the transient period
\BState \textbf{Output:} Set of benign updates $\mathcal{T}_3$
\BState \textbf{Initial:} $\mathcal{T}_3 = \emptyset$\\

\BState \textbf{Filter 3:} Temporal-similarity-based filtering
\If{$t > T_\mathsf{th}$}
\For{each $j \in [K]$}
    \State Compute $\mu_\mathrm{d}^{(t)}, \mu_\mathrm{c}^{(t)}$ as the average of the exponentially weighted time window of $W$ rounds with $\{s_j^{(t-1)}, \dots, s_j^{(t-W-1)}\}$ and $\{\beta_j^{(t-1)}, \dots, \beta_j^{(t-W-1)}\}$, respectively
    \State Compute $\sigma_\mathrm{d}^{(t)}, \sigma_\mathrm{c}^{(t)}$ as the average standard deviation of the exponentially weighted time window of $W$ rounds with $\{s_j^{(t-1)}, \dots, s_j^{(t-W-1)}\}$ and $\{\beta_j^{(t-1)}, \dots, \beta_j^{(t-W-1)}\}$, respectively\\
    \State $s_j^{(t)} = \| \theta_j^{(t)} - \theta^{(t-1)}_j \|^2$
    \State $\beta_j^{(t)} = 1 - \dfrac{\langle \theta^{(t)}_j, \theta^{(t-1)}_j  \rangle}{\| \theta^{(t)}_j \| \cdot \| \theta^{(t-1)}_j  \|}$\\  
    \If {$\mu_\mathrm{d}^{(t)} - \sigma_\mathrm{d}^{(t)}  \leq s_j^{(t)} \leq \mu_\mathrm{d}^{(t)} + \sigma_\mathrm{d}^{(t)}$ \text{ and } $\mu_\mathrm{c}^{(t)} - \sigma_\mathrm{c}^{(t)} \leq \beta_j^{(t)} \leq \mu_\mathrm{c}^{(t)} + \sigma_\mathrm{c}^{(t)}$} 
        \State $\mathcal{T}_3 = \mathcal{T}_3 \cup \{\theta_j\}$
    \EndIf
\EndFor
\EndIf
\end{algorithmic}
\end{algorithm}

The algorithm requires an initial transient period of $T_\mathsf{th}$ rounds during which it does not perform any classification. This is due to two reasons: the first and most obvious is the lack of sufficient data to determine the time window, and the second is that, due to the random and uncoordinated initialization of model parameters at each node, variations are significantly abrupt and could be mistakenly identified as attacks.

A time window of the last $W$ models of the node is taken as a reference, specifically, the last $W$ metrics of Euclidean distance and cosine similarity between the node's model updates. From these metrics, both the mean and the standard deviation are extracted by applying an Exponentially Weighted Moving Average (EWMA), i.e., the most recent metrics in the time window are considered, which is useful since patterns change over time (non-stationary). Subsequently, new metrics are calculated with the model received in the current round, and it is observed if they fall within the thresholds calculated with the mean and standard deviation. If any of them do not meet the criteria, the model update is considered abrupt and malicious in that round.

\subsection{WFAgg-E: Weighted Model Aggregation Algorithm}

The \textsf{WFAgg-E} algorithm is proposed as a model aggregation algorithm similar to the \textsf{FedAvg}. Unlike \textsf{Mean} and other variants, it performs a weighted average, giving more or less relevance to the local model of the node or to the updates from neighboring nodes. This aggregation algorithm is designed for adversarial environments so that if a malicious model is selected for aggregation, its influence is mitigated.

The fundamental idea of \textsf{WFAgg-E} consists of a variation of the first-order exponential smoother~\cite{Montgomery11} (in other fields it is called Exponential Moving Average). It started to become popular in the 1960s and is a technical indicator used in financial analysis to analyze and smooth time series data. It assigns more weight to the most recent data, making it more sensitive to recent changes.

Using the \textsf{WFAgg-E} algorithm, each client $C_i$ updates its model using its own model and the models received from its neighboring nodes, which have been previously weighted, as follows:
\begin{align}
    \notag
    {\overline{\theta}}_i &= (1-\alpha) \theta_i + \alpha \cdot \dfrac{\sum_{j \in \mathcal{N}_i(G)}{ w_{ij} \theta_j }}{\sum_{j \in \mathcal{N}_i(G)}{ w_{ij} }}\\
    &= (1-\alpha) \theta_i + \alpha \cdot \sum_{j \in \mathcal{N}_i(G)}{ w'_{ij} \theta_j},
\end{align}
where $\mathcal{N}_i(G)$ is the set of neighboring nodes of $C_i$ given the topology modeled by a graph $G$, $w_{ij}$ is the weight assigned by $C_i$ to the model of client $C_j$, $w'_{ij} \in [0,1]$ is the normalized weight (such that $\sum_{j \in \mathcal{N}_i(G)}{w'_{ij}}=1$), and $\alpha \in [0,1]$ is a smoothing factor that regulates the level of aggregation of the neighbors' models versus the current local model.

\section{Experimental setup}
\label{sec:implementation}

This section outlines the experimental setup, including the scenario, learning algorithms and dataset for image classification, primary evaluation metrics, which Byzantine-robust aggregation algorithms are evaluated and which model performance attacks are performed in the simulations.

\subsection{Validation scenario}

\hspace*{0.7em} \textbf{Topology:} We have defined a FL setting with $20$ clients, $10\%$ of them are malicious nodes. The selected bounds were chosen to ensure a balance between system stability and responsiveness to malicious behavior. Performance is evaluated in both centralized and decentralized scenarios. In the centralized scenario, a server coordinates the clients' training process, while in the decentralized scenario, each client communicates with its neighbors. The topology for the latter is modeled as an $8$-regular graph\footnote{Specifically, it is a small-world graph (also known as Watts-Strogatz), a regular graph structured in the form of a ring in which each vertex is joined to its nearest $c$ ($c$ even) neighbors. This type of graph has a random component since, with probability $p$, a new edge is added between two randomly chosen vertices. Here $p=0$ will be considered.}, i.e., each node has exactly $8$ neighbors, and each client has, at most, $25\%$ malicious neighbors. This enables a broad testing scenario to evaluate the influence of malicious nodes.

\textbf{Learning algorithm and dataset:} The nodes perform learning tasks based on image classification on the MNIST dataset \cite{LeCun10}. The MNIST dataset consists of $70,000$ handwritten digit images ($0$ to $9$), each represented as a $28 \times 28$ pixel grayscale matrix. The dataset is independently and identically distributed (IID) across all nodes, in order to isolate the effects of Byzantine attacks and the proposed aggregation strategy from those introduced by data heterogeneity (non-IID settings introduce additional convergence variability unrelated to the focus of this work). Moreover, nodes train convolutional neural networks (CNNs) as local models, with architectures consisting of 7 layers, including convolutional, pooling, and fully connected layers, among others. This architecture is similar to the LeNet-5 CNN~\cite{LeCun98}, but the specific models used can be found in the code implementation\footnote{\label{fn:code}Code is available online: \textsf{\url{https://gitlab.com/compromise3/decentralizedfedsim}}}. Each node’s training algorithm runs for 10 rounds, with the number of local iterations (or epochs) set to $1$, momentum at $0.9$, and learning rate at $\eta = 0.01$.

\textbf{Byzantine-robust aggregation algorithms:} We selected some of the most popular algorithms in the literature (detailed in Section \ref{sec:review}), along with the algorithms proposed in this paper, for the robust aggregation tests. Every node in the topology will use the same constant values and thresholds. These parameters are based on configurations commonly used in the literature on adversarial FL scenarios. For the \textsf{Trimmed-Mean} algorithm, the trim rate value is set to $\beta = 0.1$. For \textsf{Krum} and \textsf{Multi-Krum}, as well as most of the proposed algorithms (\textsf{WFAgg-D} and \textsf{WFAgg-C}), the number of malicious models is set to $f=2$ (out of $8$). Furthermore, algorithm \textsf{Multi-Krum} selects $m = K/4$ models for aggregation, i.e, a $25\%$ of the received models. For \textsf{WFAgg-T} algorithm, both the transient period and the length of the time window are set to $T_\textsf{th}=W=3$. For \textsf{WFAgg-E} algorithm, the smoothing factor is set to $\alpha = 0.8$. Finally,  for \textsf{WFAgg} algorithm, the weighting assigned to each filter are $\tau_1 = \tau_2  = 0.4, \tau_3 = 0.2$, enabling more representative scenarios for analyzing the attacks. Nonetheless, optimal weigh-filtering tuning can be guided by the type of adversarial behavior to be counteract (e.g., emphasizing distance-based metrics may enhance robustness against strong directional perturbations).

\textbf{Implementation details:} We implemented a DFL simulator using the Python programming language\footref{fn:code}. It allows the execution of both centralized and decentralized environments through the design of the network topology and the use of different classes that model the behavior of clients (in both CFL/DFL), server, etc. The ML models were designed and trained using the \texttt{PyTorch} library. To simulate multiple clients in the FL environment, we utilized the Python \texttt{threading} library, enabling concurrent training of distributed clients.

\subsection{Model Performance Attacks}

In order to assess the robustness of the different Byzantine-robust schemes, various attacks have been implemented. The attacks considered for the tests, representing the most popular poisoning attack found in the literature are detailed below \cite{Li24, Xu23} (assume that Byzantine node $C_i$ is going to perform a poisoning attack on its ML parameters model $\theta_i$):
\begin{itemize}
    \item \textbf{Noise Attack:} Byzantine nodes send modified models by injecting Gaussian noise into the benign models as $\theta_i \leftarrow \theta_i + \mathcal{N}(\mathbf{\mu}, \sigma^2 I)$, where $I \in \mathcal{M}_{d\times d}$ is the identity matrix of dimension $d$. In our experiments, both the mean and variance of the noise of each parameter is set to $0.1$, i.e.,  $\mu = 0.1 \cdot \mathbf{1} \in \mathbb{R}^d$ and $\sigma = 0.1$.
    
    \item \textbf{Sign-Flipping (SF):} The Byzantine clients send reversed models without changing their modulus as $\theta_i \leftarrow - \theta_i$. 
    
    \item \textbf{Label-Flipping (LF):} This is an example of a data poisoning attack, where Byzantine clients flip the local sample labels during the training process to generate malicious models. In particular, the label of each training sample in Byzantine clients is flipped from $\ell \in \{0, 1, \dots, C - 1\}$ to $C - 1 - \ell$, where $C$ is the total classes of labels.
    
    \item \textbf{A Little is Enough (ALIE):} This is a more sophisticated version of the Noise Attack, where model's parameters are selected carefully in order to appear benign and harm the model performance. A Little is Enough attack assumes that the benign models' parameters are expressed by a normal distribution. For each model's coordinate $j \in [d]$, the attacker computes the mean $\mu_j$ and standard deviation $\sigma_j$ over benign models and built the corresponding malicious model by taking values in range $(\mu_j - z_\text{max} \sigma_j, \mu_j + z_\text{max} \sigma_j)$, where $z_\text{max} \in [0,1]$ is typically obtained from the Cumulative Standard Normal Function. Here, the threshold parameter is set to $z_\text{max} = 0.5$.
    
    \item \textbf{Inner Product Manipulation (IPM):} The Inner Product Manipulation attack seeks for the negative inner product between the true mean of the updates and the output of the aggregation schemes. Assume that there are a set of models $\{\theta_k\}_{k=1}^{N}$ where the first $M$ are malicious updates. So, a way of performing the IPM attack is 
    \begin{align}
        \theta_1 = \cdots = \theta_M = - \dfrac{\varepsilon}{N-M} \sum_{k=M+1}^{N} \theta_k
    \end{align}
    where $\varepsilon$ is a positive coefficient controlling the magnitude of malicious updates. Then the models' mean becomes
    \begin{align}
        \dfrac{1}{N} \sum_{k \in [N]} \theta_k = \dfrac{N-M(1+\varepsilon)}{N(N-M)} \sum_{k=M+1}^{N} \theta_k.
    \end{align}
    The key of this attack lies in choosing the value of $\varepsilon$. If $\varepsilon < N/M -1$, IPM does not change the direction of the mean but decreases its magnitude, in the other case, the sign is reversed. We examine the two different cases by letting $\varepsilon = 0.5$ and $\varepsilon = 100$, respectively.
\end{itemize}

\subsection{Metrics}

The tests will primarily use two metrics: accuracy and \textsf{R-squared}. The accuracy metric (its micro average approach), which is one of the most popular metrics for a multi-class classification problems, can be defined as
\begin{align}
    \mathsf{Accuracy} = \dfrac{\sum_{\ell=0}^{C-1} \mathsf{TP}_\ell}{|\mathcal{D}_{\textsf{test}}|},
\end{align}
where $\mathsf{TP}_\ell$ are the true positives over class/label $\ell \in [C]$, $C$ is the number of classes, and $|\mathcal{D}_{\textsf{test}}|$ is the number of samples in the testset. In this work, this metric will be shown with its percentage value (\%). 

To quantify and evaluate the convergence of local models in a DFL scenario, we propose an adapted version of the well-known \textsf{R-squared} metric (also known as $R^2$) as, originally, the metric is defined for univariate cases. Given a set of $N$ multidimensional vectors $\mathbf{v}_1, \dots, \mathbf{v}_N \in \mathbb{R}^d$, we are going to define the \textsf{R-squared} metric as follows:
\begin{align}
    R^2 = 1 - \dfrac{\mathsf{SSR}}{\mathsf{SST}} = 1 - \dfrac{\sum_{i=1}^N \norm{\mathbf{v}_i - \overline{\mathbf{v}}}^2}{\sum_{i=1}^N \norm{\mathbf{v}_i}^2},
\end{align}
where:
\begin{itemize}
    \item $\mathsf{SSR}$ is the total sum of squares of the differences between each vector and the mean vector (the term to be minimized),
    \item $\mathsf{SST}$ is the total sum of squares of each vector (the term that normalizes the result of $\mathsf{SSR}$),
    \item $\overline{\mathbf{v}} = \dfrac{1}{N} \sum_{i=1}^N \mathbf{v}_i$ is the mean vector.
\end{itemize}

This metric provides a measure of the proportion of the total variability of the vectors explained by the mean vector, i.e., how similar the vectors are to each other. Applied to the DFL scenario, the set of vectors would correspond to the local models of the benign nodes, to analyze how similar these models are to each other.

\section{Results and Discussion}
\label{sec:results}

The objective of the upcoming tests, described in the following subsections, is to better understand the effectiveness of these Byzantine-robust aggregation schemes under multiple adversarial attacks. These attacks aim to degrade the model performance of the learning process, affecting both the local models accuracy and the global convergence of the learning process. Therefore, this evaluation consists of three different tests that will be analysed on the selected algorithms. First, examine and analyse the robustness of the algorithms to Byzantine attacks (Subsection \ref{sec:result1}). Then, we study the evolution of the accuracy convergence throughout the communication rounds (Subsection \ref{sec:result2}). Finally, we analyse the global consistency among benign models in decentralized scenario (Subsection \ref{sec:result3}). The same tests will be analysed for the proposed algorithms (Subsections \ref{sec:proposed1} and \ref{sec:proposed2}).

\subsection{Evaluation of state-of-the-art Byzantine-Robust Schemes}

Are the proposed Byzantine-robust aggregation algorithms for centralized scenarios equally effective when evaluated in decentralized scenarios? This section aims to answer that question by conducting various analyses on well-known Byzantine-robust aggregation algorithms from the literature for the previously mentioned metrics.

\subsubsection{General analysis of the attack effectiveness}
\label{sec:result1}
The algorithms, discussed in Section~\ref{sec:review}, that will be considered for the analysis are: \textsf{Mean}, \textsf{Trimmed-Mean}, \textsf{Median}, \textsf{Krum}, \textsf{Multi-Krum}, and \textsf{Clustering}. Table \ref{tab:accuracy_results} presents a general comparison of the robust aggregation schemes mentioned, by analyzing the accuracy metric in the testset, exposed to different poisoning attacks. Generally, most algorithms achieve good results and withstand the majority of attacks. This can be justified by the fact that the inherent simplicity of the MNIST dataset reduces the learning complexity for the employed model.

\begin{table*}[tbp]

\centering
\footnotesize

\caption{Accuracy comparing of different Byzantine-robust aggregation schemes under various attacks. ``Decentralized'' columns contain the mean accuracy of the benign nodes with 0, 1 and 2 malicious neighbors, respectively. Red highlighted values show values significantly below the average of the others.\\}
\resizebox{\textwidth}{!}{
\begin{tabular}{c | c || c | c | c | c ||| c  || c | c | c | c  } 
\hline
 \multirow{2}{*}{\textbf{Defense Scheme}} & \multirow{2}{*}{\textbf{Attack}} & \multicolumn{1}{c|}{\multirow{2}{*}{\textbf{Centralized}}} & \multicolumn{3}{c|||}{\textbf{Decentralized}}  & \multirow{2}{*}{\textbf{Defense Scheme}} & \multicolumn{1}{c|}{\multirow{2}{*}{\textbf{Centralized}}} & \multicolumn{3}{c}{\textbf{Decentralized}}     \\ \cline{4-6} \cline{9-11}
&  &   & \multicolumn{1}{c|}{0 m.n.} & \multicolumn{1}{c|}{1 m.n.} & \multicolumn{1}{c|||}{2 m.n.} &  &   & \multicolumn{1}{c|}{0 m.n.} & \multicolumn{1}{c|}{1 m.n.} & \multicolumn{1}{c}{2 m.n.} \\ \hline \hline
\multirow{7}{*}{\textsf{Mean}}  
 & No attack    & 98.08 & 97.47 & 96.90 & 94.56 &
\multirow{7}{*}{\textsf{WFAgg-D}}  
                & 97.93 &  97.33 & 97.10 & 95.00 \\
 & Noise        & \textcolor{red}{11.45} & \textcolor{red}{11.13} & \textcolor{red}{12.00} & \textcolor{red}{10.89} & 
                & 98.07 & 97.60 & 96.53 & 94.78 \\
 & SF           & 97.00 & 96.87 & 95.77 & 92.11  &
                & 98.04 & 97.73 & 96.90 & 94.78 \\
 & LF           & 97.10 &  97.47 & 97.07 & 92.33 &
                & 97.99 & 97.33 & 96.87 & 94.67 \\
 & IPM-0.5      & 97.22 & 96.87 & 96.07 & 93.44 &
                & 98.10 &  97.40 & 96.80 & 93.56\\
 & IPM-100      & \textcolor{red}{9.80} & \textcolor{red}{10.87} & \textcolor{red}{10.50} & \textcolor{red}{10.11} &
                & 97.71 & 97.87 & 97.07 & 93.67 \\
 & ALIE         & 98.00 & 97.60 & 96.93 & 94.11 &
                & 98.04 &  98.00 & 96.67 & 95.22 \\ \hline \hline
 \multirow{7}{*}{\textsf{Trimmed-Mean}}  
 & No attack    & 98.0 &  97.33 & 96.77 & 95.44 &
 \multirow{7}{*}{\textsf{WFAgg-C}}  
                & 97.98 & 97.87 & 96.90 & 95.11 \\
 & Noise        & 98.15 & \textcolor{red}{11.13} & \textcolor{red}{11.97} & \textcolor{red}{10.89} &
                & 98.06 & 97.87 & 77.97 & \textcolor{red}{14.22} \\
 & SF           & 97.89 & 96.80 & 95.70 & 92.00 &
                & 97.77 & 97.67 & 96.30 & 94.67 \\
 & LF           & 98.08 & 97.07 & 95.83 & 85.11 &
                & 98.02 & 97.53 & 96.80 & 94.89 \\ 
 & IPM-0.5      & 97.98 & 96.87 & 96.30 & 94.11 &
                & 98.14 & 97.40 & 96.90 & 95.00 \\
 & IPM-100      & 97.73 & \textcolor{red}{10.87} & \textcolor{red}{10.50} & \textcolor{red}{10.11} &
                & 97.89 & 97.80 & 96.63 & 94.44 \\
 & ALIE         & 97.93 & 97.73 & 96.83 & 95.78 &
                & 98.00 & 97.40 & 97.10 & 95.44 \\ \hline \hline
 \multirow{7}{*}{\textsf{Median}}  
 & No attack    & 98.0 & 97.13 & 96.73 & 94.89 &
 \multirow{7}{*}{\textsf{WFAgg-T}}  
                & 97.82 & 97.00 & 96.30 & 94.44 \\
 & Noise        & 98.01 & 97.13 & 96.73 & 93.00 &
                & \textcolor{red}{11.35}  & \textcolor{red}{11.13} & \textcolor{red}{11.57} & \textcolor{red}{10.22} \\
 & SF           & 97.88 & 97.33 & 96.733 & 92.89 &
                & 96.94 & 96.73 & 95.40 & 93.67 \\
 & LF           & 98.07 & 96.67 & 96.53 & 93.33 &
                & 97.50 & 96.53 & 96.13 & 94.22 \\
 & IPM-0.5      & 97.99 & 97.73 & 95.70 & 89.22 &
                & \textcolor{red}{11.35} & 97.40 & 95.97 & 94.22 \\
 & IPM-100      & 98.16 & 97.13 & 96.33 & 92.00 &
                & \textcolor{red}{9.8} & \textcolor{red}{10.87} & \textcolor{red}{10.50} & \textcolor{red}{10.11} \\
 & ALIE         & 97.95 & 97.07 & 96.67 & 92.33 &
                & 98.59 & 97.47 & 96.80 & 94.56 \\ \hline \hline
 \multirow{7}{*}{\textsf{Krum}}  
 & No attack    & 96.71 & 97.00 & 96.47 & 93.78 &
 \multirow{7}{*}{\textsf{WFAgg-E}}  
                & 97.72 &  97.67 & 97.10 & 94.56\\
 & Noise        & 97.08 & 96.40 & 96.23 & 95.56 &
                & \textcolor{red}{12.35} & \textcolor{red}{11.53} & \textcolor{red}{11.13} & \textcolor{red}{10.44} \\
 & SF           & 95.42 & 96.67 & 96.67 & 94.33 &
                & 96.57 & 97.47 & 96.03 & 92.78 \\
 & LF           & 95.53 & 95.93 & 95.53 & 94.22 &
                & 97.46 & 97.33 & 96.90 & 94.78 \\
 & IPM-0.5      & 96.12 & 96.07 & \textcolor{red}{11.57} & \textcolor{red}{10.22} &
                & 97.24 & 97.00 & 96.10 & 93.11 \\
 & IPM-100      & 95.87 & 97.47 & 95.83 & 94.56 &
                & \textcolor{red}{10.87} & \textcolor{red}{10.87} & \textcolor{red}{10.50} & \textcolor{red}{10.11} \\
 & ALIE         & 98.19 & 97.60 & 96.93 & 96.11 &
                & 97.90 & 97.67 & 96.57 & 95.56 \\ \hline \hline
 \multirow{7}{*}{\textsf{Multi-Krum}}  
 & No attack    & 97.71 & 96.33 & 97.0 & 94.44 & 
 \multirow{7}{*}{\textsf{Alt-WFAgg}}  
                & 98.46 &  97.27 & 97.10 & 95.67\\
 & Noise        & 97.66 & 97.07 & 96.67 & 92.00 &
                & 98.05 & 97.40 & 96.73 & \textcolor{red}{56.78} \\
 & SF           & 97.92 & 97.07 & 97.17 & 94.33 &
                & 98.20 &  97.33 & 96.87 & 94.56\\
 & LF           & 97.97 & 97.53 & 96.97 & 95.11 &
                & 98.26 & 97.33 & 96.57 & 92.78 \\
 & IPM-0.5      & 97.7 & 97.40 & 95.33 & 87.11 &
                & 98.21 & 97.67 & 96.90 & 95.44 \\
 & IPM-100      & 97.83 & 97.87 & 96.93 & 95.11 &
                & 98.24 & 97.40 & 96.73 & 94.67 \\
 & ALIE         & 97.76 & 97.33 & 97.23 & 95.33 &
                & 98.09 & 97.13 & 96.83 & 95.33 \\ \hline \hline
 \multirow{7}{*}{\textsf{Clustering}}  
 & No attack    & 98.07 & 97.60 & 96.73 & 94.44 &
 \multirow{7}{*}{\textsf{WFAgg}}  
                & 98.61 & 97.40 & 96.40 & 95.33 \\
 & Noise        & 96.86 & 97.53 & \textcolor{red}{10.44} & \textcolor{red}{9.93} &
                & 98.44 & 97.53 & 96.57 & 94.89 \\
 & SF           & 98.14 & 97.73 & 96.63 & 95.56 &
                & 98.26 & 97.53 & 97.13 & 95.56 \\
 & LF           & 98.14 & 97.87 & 95.30 & 90.67 &
                & 98.53 & 97.20 & 96.93 & 95.78 \\
 & IPM-0.5      & 97.98 & 97.53 & 96.43 & 95.44 &
                & 98.13 &  97.20 & 96.70 & 95.22\\
 & IPM-100      & 98.16 & 97.47 & 96.57 & 95.22 &
                & 98.35 & 96.80 & 96.93 & 94.78 \\
 & ALIE         & 98.06 & 97.20 & 96.77 & 95.56 &
                & 98.20 & 97.27 & 97.03 & 95.44 \\ \hline
\end{tabular}
}
\label{tab:accuracy_results}
\end{table*}

However, some of the deployed attacks demonstrate greater effectiveness in preventing model convergence, being capable of compromising the robustness of several algorithms. Among these attacks are Noise and IPM-100, both characterized by drastically manipulating the magnitude of malicious models, capable of reducing accuracy to minimal values. These attacks can prevent the convergence of accuracy for both \textsf{Mean} and \textsf{Trimmed-Mean} algorithms. The former is particularly notable since it can not converge in the centralized scenario, which does not occur with the \textsf{Trimmed-Mean} algorithm due to the filtering of adverse values. The \textsf{Clustering} algorithm is also affected by the Noise attack but not in the same way as the previous ones. In contrast, only nodes with direct malicious neighbors seem to be affected, while the rest do not appear to experience negative effects.

Other attacks, such as IPM-0.5 or Sign Flipping, do not appear to significantly worsen the results compared to the scenario without attacks. This is because they focus not on drastically manipulating the magnitude of malicious models but on changing the gradient that optimizes the learning algorithm in the opposite direction. This is not the case for the \textsf{Krum} algorithm, which is profoundly affected by the IPM-0.5 attack. Since it is an attack with a magnitude that does not stand out compared to benign neighbors, the algorithm does not seem capable of distinguishing it in Euclidean distance from the rest. Even so, some algorithms like \textsf{Multi-Krum} show a deterioration in results without preventing the convergence of accuracy (e.g., in the worst-case scenario, $87.11$\% is obtained for IPM-0.5). 

Data poisoning attacks, such as Label Flipping, do not appear to cause any degradation in accuracy across various algorithms (e.g., in the worst-case scenario, obtaining an accuracy of $85.11$\% and $90.67$\% for \textsf{Trimmed-Mean} and \textsf{Clustering}, respectively). Lastly, the ALIE attack does not seem to compromise the algorithm's convergence in any case, as the potential of this attack lies in the high variance of benign models --- a situation that does not occur with the employed dataset.

\begin{figure}[tbp]
    \centering
    \includegraphics[width=1.0\textwidth]{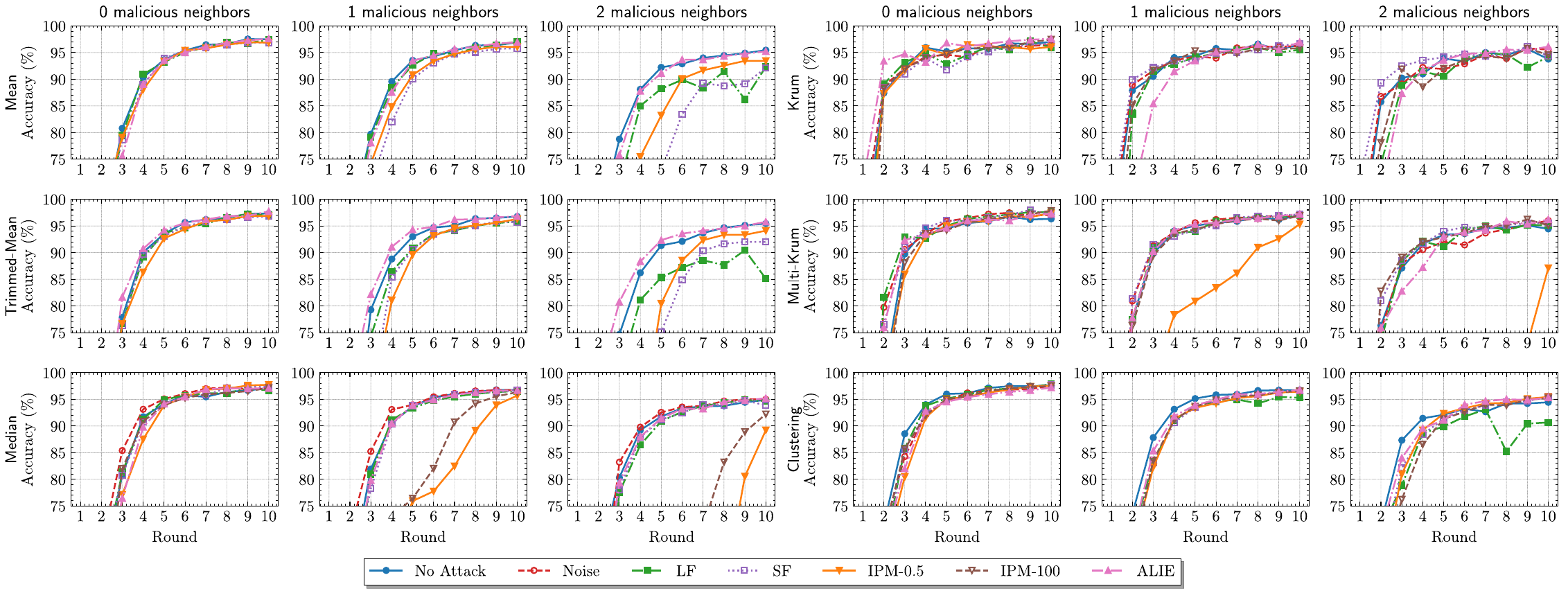}
    \caption{Comparison of average accuracy in a decentralized scenario using state-of-the-art Byzantine-robust aggregation algorithms, depending on the number of malicious neighbors (by columns, from left to right: \textsf{Mean}, \textsf{Trimmed-Mean}, \textsf{Median}, \textsf{Krum}, \textsf{Multi-Krum}, \textsf{Clustering}).} 
    \label{fig:sota-decentralized}
\end{figure}

\subsubsection{Robustness analysis of aggregation algorithms}
\label{sec:result2}
In any case, a detailed study of each of the Byzantine-robust aggregation algorithms is required. To this end, Figure \ref{fig:sota-decentralized} displays the evolution of accuracy under the corresponding testset in a decentralized scenario. Due to the disparity of results and the multiple scenarios that can be found in decentralized environments, Figure \ref{fig:sota-decentralized} shows the average accuracy of the nodes based on the number of their malicious neighbors (this is the main factor that can influence the experiment's outcome).

Algorithms like \textsf{Mean}, \textsf{Trimmed-Mean}, or \textsf{Median}, categorized as central tendency statistics-based algorithms, do not start with the best results in this study. The results using the \textsf{Mean} algorithm will be taken as a reference for the rest of the analysis because its outcomes are one of the worst compared to the others (in general terms). In fact, it is the only algorithm that does not allow convergence in the centralized scenario under Noise and IPM-100 attacks. Moreover, focusing on the decentralized scenario, it is observed that in most of the attacks, almost all simulated rounds are required to achieve convergence of accuracy (specifically, Label-Flipping, Sign-Flipping, and IPM-0.5).

The drawback in centralized scenarios is partially remedied under \textsf{Trimmed-Mean}, but the results do not improve in the decentralized scenario due to the reduced number of models to aggregate by each node. However, the \textsf{Median} algorithm can withstand most attacks without deterioration in accuracy by considering only the central values of the parameter models, although this process is not constant throughout the communication rounds and is not always capable of mitigating designed attacks like IPM-0.5 and IPM-100 in decentralized scenarios. In both cases, \textsf{Trimmed-Mean} and \textsf{Median} algorithms, similar to \textsf{Mean}, the convergence of accuracy in decentralized scenarios is relatively slow compared to other algorithms.

On the other hand, Euclidean distance-based algorithms like \textsf{Krum} and \textsf{Multi-Krum} are able to resist attacks such as Noise and IPM-100 since their goal is to detect malicious models that are far from the rest in Euclidean distance. This is a drawback in itself as attacks like IPM-0.5 are not easily detectable with these type of measures, as observed with \textsf{Krum} in a decentralized scenario. In both configurations of the scenario, Krum does not exhibit uniformity in the evolution of accuracy in most attacks because it selects only one model (the one that achieves the best metric), leading to a loss of information.

These attacks and drawbacks are mitigated in \textsf{Multi-Krum} due to it considers several neighbor models (the closest to the rest) in the aggregation process. Even so, the IPM-0.5 attack does not seem to be easily detectable by the algorithm in decentralized scenarios, causing a slowdown in the convergence of accuracy. In any case, Euclidean distance-based algorithms show a deterioration in accuracy, especially when the number of malicious nodes present in the scenario is not correctly estimated. Despite this, it is worth noting good results in \textsf{Multi-Krum} for decentralized scenarios with a smaller number of models received.

\textsf{Clustering}, as a representation of cosine distance-based algorithms, fails in the Noise attack in the decentralized scenario. This is likely because this attack does not produce a drastic change in the model's direction but in its magnitude, so the cosine distances involving the malicious model do not vary much. Due to the use of hierarchical clustering algorithms, having fewer models in the decentralized scenario allows including the model in a smaller cluster.

\subsubsection{Model consistency analysis in decentralized scenarios}
\label{sec:result3}

\begin{figure}[tbp]
   \centering
   \includegraphics[width=0.55\textwidth]{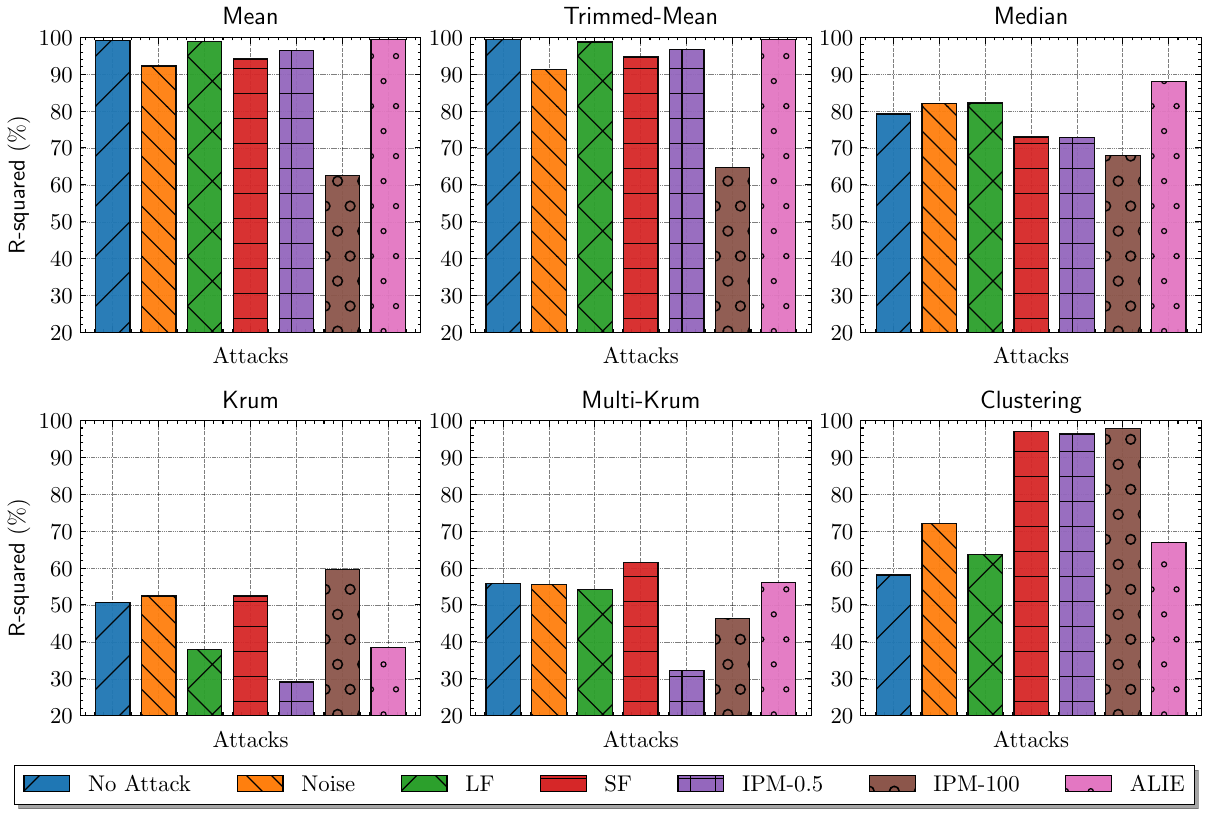}
   \caption{Comparison of \textsf{R-Squared}, $R^2$, metric (last round of federation) in a decentralized scenario using state-of-the-art Byzantine-robust aggregation algorithms.}
   \label{fig:r-squared-sota}
\end{figure}

To study the convergence of models among different benign nodes in a decentralized scenario, several experiments have been conducted using the \textsf{R-Squared} ($R^2$) metric defined in Section \ref{sec:implementation}. The results can be observed in Figure \ref{fig:r-squared-sota}.

In algorithms such as \textsf{Mean} and \textsf{Trimmed-Mean}, convergence is unquestionable because the $R^2$ metric is based on the average of the benign nodes, and these algorithms are based on central tendency statistics. Nodes compute the mean with the models they receive from their neighbors, and these from theirs, facilitating convergence. In the case of \textsf{Median}, convergence is slower and even insufficient under some attacks because it takes the central parameters of the models, and each node has received different models (probably that ``central'' value will differ).

The situation is different for \textsf{Krum} and \textsf{Multi-Krum}, where convergence does not reach $60\%$ in practically any of the attacks. This is the main disadvantage of Byzantine-robust aggregation algorithms that perform a filtering of models, as each network node selects a different set of models based on those received, which does not facilitate convergence.

Similarly, the $R^2$ results under the \textsf{Clustering} scheme can be explained as well, where certain attacks (Noise, ALIE, Label Flipping, \dots) prevent a progressive evolution of convergence. In this case, the filtering is done through a hierarchical clustering algorithm based on cosine distances, causing the cluster selected by each node to differ among them.

\subsection{Evaluation of Proposed Byzantine-Robust Schemes}

Once the Byzantine-robust aggregation algorithms found in the literature on decentralized scenarios have been evaluated, it can be confirmed that none of them simultaneously excels in all three analyses conducted. Among them, the \textsf{Multi-Krum} algorithm stands out as it is capable of practically mitigating the employed attacks, although it does not achieve good convergence among models. Now, the same analysis will be conducted for the algorithms proposed in this document.

\subsubsection{Performance analysis of proposed filtering algorithms}
\label{sec:proposed1}
\begin{figure}[tbp]
   \centering
   \includegraphics[width=1.0\textwidth]{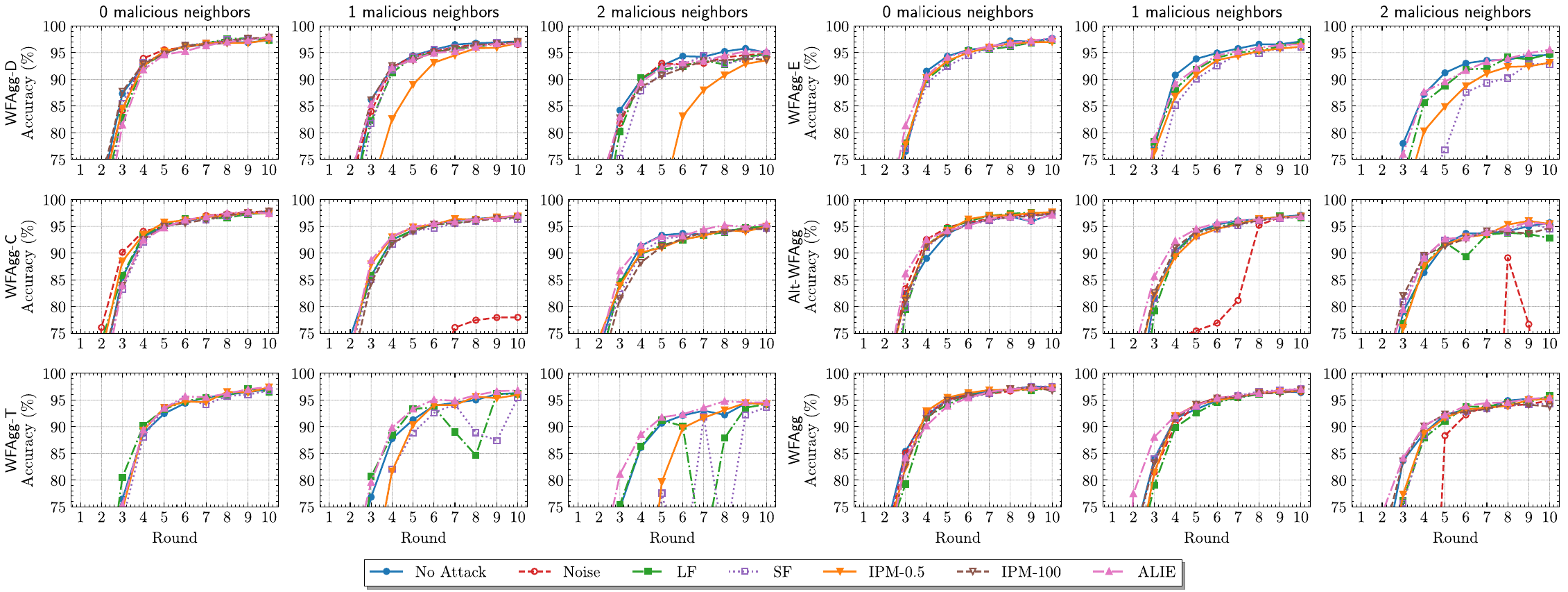}
   \caption{Aggregated accuracy evolution in a decentralized scenario using the different proposed Byzantine-robust algorithms (by columns, from left to right, \textsf{WFAgg-D}, \textsf{WFAgg-C}, \textsf{WFAgg-T}, \textsf{WFAgg-E}, \textsf{Alt-WFAgg}, \textsf{WFAgg}).} 
   \label{fig:proposal-decentralized}
\end{figure}

This section aims to evaluate and compare the single proposed algorithms of this paper in order to highlight their improvements and drawbacks compared to the state-of-the-art ones. Those algorithms are: \textsf{WFAgg-D}, \textsf{WFAgg-C}, \textsf{WFAgg-T} and \textsf{WFAgg-E}. Both Table \ref{tab:accuracy_results} and Figure \ref{fig:proposal-decentralized} displays the results of experiments conducted in both centralized and decentralized scenarios.

First, the \textsf{WFAgg-D} and \textsf{WFAgg-C} algorithms will be evaluated. These Byzantine-robust aggregation algorithms are based on Euclidean distances and cosine distances, respectively. It can be observed that both algorithms are capable of mitigating all evaluated attacks, except for \textsf{WFAgg-C} under the Noise attack, where the accuracy is affected (although only in the worst-case scenario, as it does converge in less adverse scenarios). In both cases, the convergence of accuracy occurs within a few communication rounds. Compared to the most similar state-of-the-art algorithms, both algorithms achieve better results in the majority of attacks against \textsf{Multi-Krum} and \textsf{Clustering}, respectively. This can be explained by the use of the median model as a reference due to its statistical properties and the benefits of Euclidean/cosine distances for mitigating attacks.

Both algorithms stand out compared to their counterparts, especially in the previously described weaknesses. Under the IPM-0.5 attack, both the accuracy convergence and final result of \textsf{WFAgg-D} clearly improve over those of \textsf{Multi-Krum}. Under the Noise attack, \textsf{WFAgg-C} outperforms \textsf{Clustering}, being capable of completely mitigating the attacks in some decentralized scenarios, and, when not entirely mitigating, still improving the results.

Regarding the results of the \textsf{WFAgg-E} aggregation algorithm, it is observed that it is not capable of mitigating the same attacks as in the case of the \textsf{Mean} or \textsf{Trimmed-Mean} algorithm, since it is ultimately a type of weighted average. However, some results improve slightly in adverse scenarios (such as decentralized nodes with $2$ malicious neighbors).

It does not make sense to analyze the results of the \textsf{WFAgg-T} algorithm considering it as a Byzantine-robust aggregation algorithm. This seeks to detect temporal anomalies between rounds of models sent by nodes, so that if a node attacks in the same way consistently throughout the rounds, it is not considered an attack. Despite this, the algorithm is capable of mitigating several attacks, and in those where convergence is not possible (IPM-100 or Noise), it can be justified by the controlled modification of the model's magnitude.

\begin{figure}[tbp]
   \centering
   \includegraphics[width=0.55\textwidth]{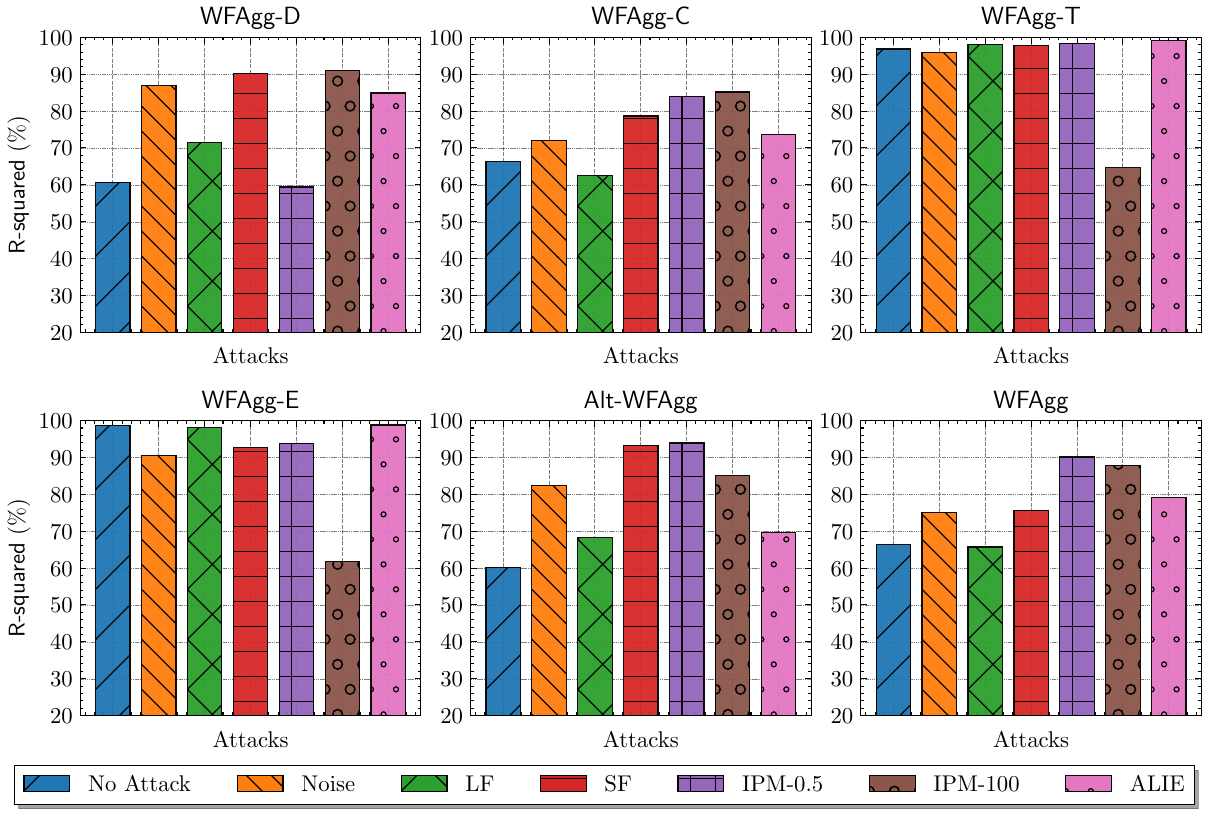}
   \caption{Comparison of \textsf{R-Squared}, $R^2$, metric (last round of federation) in a decentralized scenario using the proposed Byzantine-robust algorithms.} 
   \label{fig:R_squared-proposal}
\end{figure}

Regarding the convergence of models with \textsf{R-Squared} in decentralized scenarios, see Figure \ref{fig:R_squared-proposal}. Algorithms like \textsf{WFAgg-E} or \textsf{WFAgg-T} converge without major difficulty because the aggregation of models is based on averages and no filtering is performed, as detailed earlier. In general terms, the convergence of \textsf{WFAgg-D} and \textsf{WFAgg-C} improves those of \textsf{Multi-Krum} (notably) and \textsf{Clustering}, respectively. It is worth noting that in the latter case, the Noise attack slows down convergence as occurs with \textsf{Clustering} due to the use of cosine distances and clustering techniques.

\subsubsection{Performance analysis of proposed \textsf{WFAgg}} 
\label{sec:proposed2}

Finally, this section studies and analyzes the results of the work proposal, an algorithm based on weighted filtering of malicious models. Two proposals are evaluated: \textsf{WFAgg} which uses the algorithms proposed in this work based on distance measures (\textsf{WFAgg-D} and \textsf{WFAgg-C}) and \textsf{Alt-WFAgg}, an alternative version of the previous algorithm, which employs some state-of-the-art algorithms based on the same techniques (\textsf{Multi-Krum} and \textsf{Clustering}). The results are also presented in Table \ref{tab:accuracy_results} and Figures \ref{fig:proposal-decentralized}, and \ref{fig:R_squared-proposal}, in the same way as before.

\begin{figure}[tbp]
   \centering
   \includegraphics[width=1.0\textwidth]{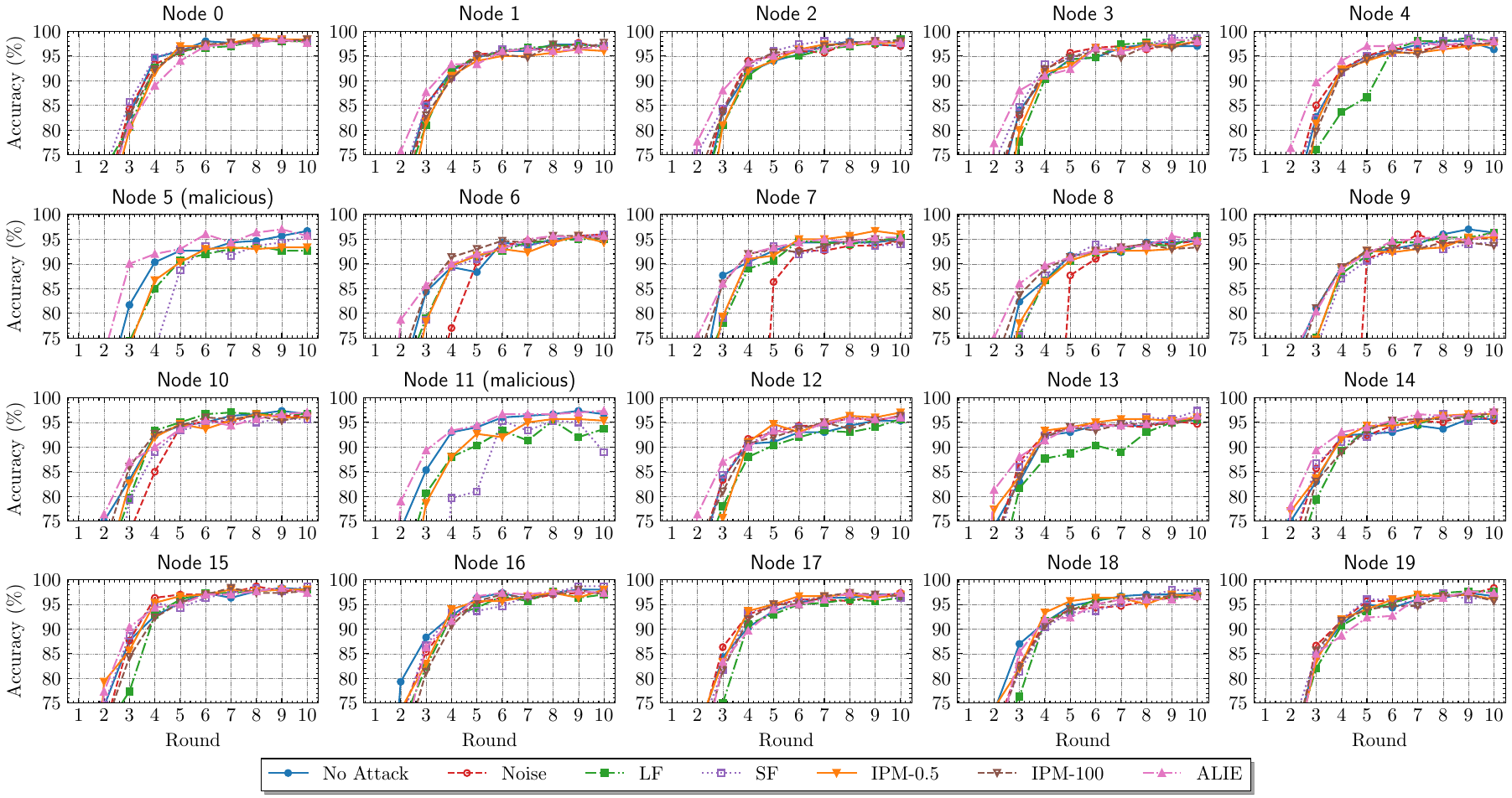}
   \caption{Accuracy evolution in a decentralized scenario using \textsf{WFAgg} Byzantine-robust algorithm (nodes 5 and 11 are malicious)} 
   \label{fig:all-nodes-proposal}
\end{figure}

In general terms, the results in a centralized environment are correct in line with the majority of previously evaluated scenarios. Focusing on decentralized scenarios, the \textsf{Alt-WFAgg} algorithm achieves good results, only compromising its performance under the Noise attack. This is mainly due to the influence of the \textsf{Clustering} algorithm (which was unable to converge under this attack). However, in the case where there is only one malicious neighbor, it requires a greater number of rounds to converge but the final result is correct. In the worst-case scenario with two malicious neighbors, accuracy convergence is not reached, but the final result is better, $46.78\%$. Regarding model convergence in $R^2$, the results vary depending on the type of attack, although they are better than in some previous cases.

Finally, the \textsf{WFAgg} algorithm achieves similar but slightly better results than the previous ones. It stands out mainly for being able to mitigate all the developed attacks, both in the final accuracy result and the accuracy convergence evolution. Figure \ref{fig:proposal-decentralized} shows how the \textsf{WFAgg} algorithm is capable of mitigating attacks from the initial learning rounds. Particularly noteworthy are the good results in the worst proposed scenario (nodes in a decentralized setting with two malicious neighbors), where the Noise attack is filtered from round $5$ (a fact not met by the previous algorithm). Regarding model convergence in $\mathbb{R}^2$, the results are similar to those obtained with the \textsf{Alt-WFAgg} algorithm. However, our method shows a slight improvement both in the absence of attacks and under strong attack scenarios such as IMP-100 (one of the most effective in previous results) and ALIE. In general terms, \textsf{WFAgg} obtains the best results if we only consider algorithms capable of mitigating the majority of attacks (which correspond with those that apply some type of filtering, in most cases).

To observe the detailed behavior of the \textsf{WFAgg} algorithm, experiments have been conducted to monitor the accuracy evolution of each node separately. The results are shown in Figure \ref{fig:all-nodes-proposal}, where it can be seen that all network nodes, excluding the malicious ones, converge in accuracy without difficulty in just a few rounds under all the proposed attacks.

\section{Conclusions and Future Work}
\label{sec:conclusions}

This proposal presents an algorithmic solution to strengthen security against malicious attacks in Decentralized Federated Learning environments. It integrates techniques based on Byzantine-robust algorithms, which mitigate security vulnerabilities in Federated Learning, particularly when central servers are removed to achieve true decentralization. The integration of various techniques, such as security filters based on distance, similarity, and temporal changes, enables the identification and mitigation of malicious contributions from Byzantine nodes and adapts to the adverse conditions posed by the variability of decentralized scenarios.

Experiments demonstrated that the proposed algorithm effectively maintains model accuracy and achieves satisfactory model convergence in the presence of all the considered Byzantine attacks, outperforming multiple centralized algorithm proposals evaluated in decentralized scenarios. The combination of multiple filters allowed for more precise detection of malicious contributions by weighting the results of these filters and considering different decisions for a potential Byzantine attack. Therefore, the integration of Byzantine-robust schemes in Decentralized Federated Learning scenarios establishes an efficient alternative to address the challenges faced by such environments.

However, further studies are still needed in the field of Decentralized Federated Learning. Continuing with the proposed work, it is interesting to study in depth the optimization of communication networks to reduce overhead and latency, among other factors, to enhance overall learning process performance. In this regard, we are working on the implementation of technologies such as Software-Defined Networks (SDN) to not only improve network parameters as mentioned but also to propose scalable and dynamic DFL environments. This paradigm allows exploration of other alternatives, such as enhancing the security of the learning process from the perspective of communication between devices.

\section*{Acknowledgment}

This work was supported by the grant PID2020-113795RB-C33
funded by MICIU/AEI/10.13039/501100011033 (COMPROMISE project), the grant PID2023-148716OB-C31 funded by MCIU/AEI/10.13039/501100011033 (DISCOVERY project); and ``TRUFFLES: TRUsted Framework for Federated LEarning Systems'', within the strategic cybersecurity projects (INCIBE, Spain), funded by the Recovery, Transformation and Resilience Plan (European Union, Next Generation). Additionally, it also has been funded by the Galician Regional Government under project ED431B 2024/41 (GPC).



\begin{thebibliography}{10}
\providecommand{\url}[1]{#1}
\csname url@samestyle\endcsname
\providecommand{\newblock}{\relax}
\providecommand{\bibinfo}[2]{#2}
\providecommand{\BIBentrySTDinterwordspacing}{\spaceskip=0pt\relax}
\providecommand{\BIBentryALTinterwordstretchfactor}{4}
\providecommand{\BIBentryALTinterwordspacing}{\spaceskip=\fontdimen2\font plus
\BIBentryALTinterwordstretchfactor\fontdimen3\font minus \fontdimen4\font\relax}
\providecommand{\BIBforeignlanguage}[2]{{%
\expandafter\ifx\csname l@#1\endcsname\relax
\typeout{** WARNING: IEEEtran.bst: No hyphenation pattern has been}%
\typeout{** loaded for the language `#1'. Using the pattern for}%
\typeout{** the default language instead.}%
\else
\language=\csname l@#1\endcsname
\fi
#2}}
\providecommand{\BIBdecl}{\relax}
\BIBdecl

\bibitem{Mbock20}
\BIBentryALTinterwordspacing
M.~M. Ogonji, G.~Okeyo, and J.~M. Wafula, ``A survey on privacy and security of internet of things,'' \emph{Computer Science Review}, vol.~38, p. 100312, 2020. [Online]. Available: \url{https://www.sciencedirect.com/science/article/pii/S1574013720304123}
\BIBentrySTDinterwordspacing

\bibitem{Suganyadevi22}
S.~Suganyadevi, V.~Seethalakshmi, and K.~Balasamy, ``A review on deep learning in medical image analysis,'' \emph{International Journal of Multimedia Information Retrieval}, vol.~11, no.~1, pp. 19--38, 2022.

\bibitem{Shokri15}
\BIBentryALTinterwordspacing
R.~Shokri and V.~Shmatikov, ``Privacy-preserving deep learning,'' in \emph{Proceedings of the 22nd ACM SIGSAC Conference on Computer and Communications Security}, ser. CCS '15.\hskip 1em plus 0.5em minus 0.4em\relax New York, NY, USA: Association for Computing Machinery, 2015, p. 1310–1321. [Online]. Available: \url{https://doi.org/10.1145/2810103.2813687}
\BIBentrySTDinterwordspacing

\bibitem{Liu20}
\BIBentryALTinterwordspacing
B.~Liu, M.~Ding, S.~Shaham, W.~Rahayu, F.~Farokhi, and Z.~Lin, ``When {Machine} {Learning} {Meets} {Privacy}: {A} {Survey} and {Outlook},'' 2020, version Number: 1. [Online]. Available: \url{https://arxiv.org/abs/2011.11819}
\BIBentrySTDinterwordspacing

\bibitem{McMahan17}
H.~B. McMahan, E.~Moore, D.~Ramage, S.~Hampson, and B.~A. y~Arcas, ``Communication-efficient learning of deep networks from decentralized data,'' \emph{Proc. of the 20th Int. Conf. on Artificial Intelligence and Statistics (AISTATS) 2017. JMLR: W\&CP volume 54}, 2016, doi: 10.48550/arXiv.1602.05629.

\bibitem{Bonawitz17}
K.~Bonawitz, V.~Ivanov, B.~Kreuter, A.~Marcedone, H.~B. McMahan, S.~Patel, D.~Ramage, A.~Segal, and K.~Seth, ``Practical secure aggregation for privacy-preserving machine learning,'' in \emph{Proc. of the 2017 ACM SIGSAC Conf. on Computer and Communications Security}, ser. CCS ’17.\hskip 1em plus 0.5em minus 0.4em\relax ACM, Oct. 2017, doi: 10.1145/3133956.3133982.

\bibitem{Truex19}
\BIBentryALTinterwordspacing
S.~Truex, N.~Baracaldo, A.~Anwar, T.~Steinke, H.~Ludwig, R.~Zhang, and Y.~Zhou, ``A hybrid approach to privacy-preserving federated learning,'' 2019. [Online]. Available: \url{https://arxiv.org/abs/1812.03224}
\BIBentrySTDinterwordspacing

\bibitem{Yuan24}
\BIBentryALTinterwordspacing
L.~Yuan, Z.~Wang, L.~Sun, P.~S. Yu, and C.~G. Brinton, ``Decentralized federated learning: A survey and perspective,'' 2024. [Online]. Available: \url{https://arxiv.org/abs/2306.01603}
\BIBentrySTDinterwordspacing

\bibitem{Li22}
Q.~Li, B.~Kailkhura, R.~Goldhahn, P.~Ray, and P.~K. Varshney, ``Robust decentralized learning using admm with unreliable agents,'' \emph{IEEE Trans. on Signal Processing}, vol.~70, pp. 2743--2757, 2022.

\bibitem{Lyu20}
L.~Lyu, H.~Yu, and Q.~Yang, ``Threats to federated learning: A survey,'' Mar. 2020, doi: 10.48550/ARXIV.2003.02133.

\bibitem{Fang21}
M.~Fang, X.~Cao, J.~Jia, and N.~Z. Gong, ``Local model poisoning attacks to byzantine-robust federated learning,'' Nov. 2021, doi: 10.48550/ARXIV.1911.11815.

\bibitem{Marano09}
S.~Marano, V.~Matta, and L.~Tong, ``Distributed detection in the presence of byzantine attacks,'' \emph{IEEE Trans. on Signal Processing}, vol.~57, no.~1, pp. 16--29, 2009.

\bibitem{Pandi23}
\BIBentryALTinterwordspacing
V.~P. Chellapandi, A.~Upadhyay, A.~Hashemi, and S.~H.~. Zak, ``On the convergence of decentralized federated learning under imperfect information sharing,'' 2023. [Online]. Available: \url{https://arxiv.org/abs/2303.10695}
\BIBentrySTDinterwordspacing

\bibitem{Mothukuri21}
\BIBentryALTinterwordspacing
V.~Mothukuri, R.~M. Parizi, S.~Pouriyeh, Y.~Huang, A.~Dehghantanha, and G.~Srivastava, ``A survey on security and privacy of federated learning,'' \emph{Future Generation Computer Systems}, vol. 115, pp. 619--640, 2021. [Online]. Available: \url{https://www.sciencedirect.com/science/article/pii/S0167739X20329848}
\BIBentrySTDinterwordspacing

\bibitem{Blanchard17}
P.~Blanchard, E.~M.~E. Mhamdi, R.~Guerraoui, and J.~Stainer, ``Byzantine-tolerant machine learning,'' 2017, doi: 10.48550/arXiv.1703.02757.

\bibitem{Pillutla22}
K.~Pillutla, S.~M. Kakade, and Z.~Harchaoui, ``Robust aggregation for federated learning,'' \emph{IEEE Trans. on Signal Processing}, vol.~70, pp. 1142--1154, 2022.

\bibitem{Li21}
Z.~Li, L.~Liu, J.~Zhang, and J.~Liu, ``Byzantine-robust federated learning through spatial-temporal analysis of local model updates,'' 2021, doi: 10.48550/arXiv.2107.01477.

\bibitem{Kairouz21}
P.~Kairouz, H.~B. McMahan, B.~Avent, A.~Bellet, M.~Bennis, A.~N. Bhagoji, K.~Bonawitz, Z.~Charles, G.~Cormode, R.~Cummings, R.~G.~L. D'Oliveira, H.~Eichner, S.~E. Rouayheb, D.~Evans, J.~Gardner, Z.~Garrett, A.~Gascón, B.~Ghazi, P.~B. Gibbons, M.~Gruteser, Z.~Harchaoui, C.~He, L.~He, Z.~Huo, B.~Hutchinson, J.~Hsu, M.~Jaggi, T.~Javidi, G.~Joshi, M.~Khodak, J.~Konečný, A.~Korolova, F.~Koushanfar, S.~Koyejo, T.~Lepoint, Y.~Liu, P.~Mittal, M.~Mohri, R.~Nock, A.~Özgür, R.~Pagh, M.~Raykova, H.~Qi, D.~Ramage, R.~Raskar, D.~Song, W.~Song, S.~U. Stich, Z.~Sun, A.~T. Suresh, F.~Tramèr, P.~Vepakomma, J.~Wang, L.~Xiong, Z.~Xu, Q.~Yang, F.~X. Yu, H.~Yu, and S.~Zhao, ``Advances and open problems in federated learning,'' Norwell, MA, 2021, doi: 10.48550/arXiv.1912.04977.

\bibitem{He18}
\BIBentryALTinterwordspacing
L.~He, A.~Bian, and M.~Jaggi, ``{COLA}: {Decentralized} {Linear} {Learning},'' 2018, version Number: 4. [Online]. Available: \url{https://arxiv.org/abs/1808.04883}
\BIBentrySTDinterwordspacing

\bibitem{Geiping20}
J.~Geiping, H.~Bauermeister, H.~Dröge, and M.~Moeller, ``Inverting gradients -- how easy is it to break privacy in federated learning?'' 2020, doi: 10.48550/arXiv.2003.14053.

\bibitem{Beltran23}
\BIBentryALTinterwordspacing
E.~T. Martínez~Beltrán, M.~Q. Pérez, P.~M.~S. Sánchez, S.~L. Bernal, G.~Bovet, M.~G. Pérez, G.~M. Pérez, and A.~H. Celdrán, ``Decentralized federated learning: Fundamentals, state of the art, frameworks, trends, and challenges,'' \emph{IEEE Commun. Surveys and Tutorials}, vol.~25, no.~4, p. 2983–3013, 2023, doi: 10.1109/comst.2023.3315746. [Online]. Available: \url{http://dx.doi.org/10.1109/COMST.2023.3315746}
\BIBentrySTDinterwordspacing

\bibitem{Ormandi13}
R.~Ormándi, I.~Hegedüs, and M.~Jelasity, ``Gossip learning with linear models on fully distributed data,'' 2011, doi: 10.1002/cpe.2858.

\bibitem{Prabhakar22}
N.~Prabhakar and S.~Kaul, ``Decentralized federated learning-solutions based on gossip protocol and blockchain,'' \emph{20th SC@ RUG 2022-2023}, p.~92, 2022.

\bibitem{Abdelghany22}
B.-E.~A. Abdelghany, M.~Fernandez-Veiga, A.~Fernandez-Vilas, A.~M. Hassan, W.~M. Abdelmoez, and N.~El-Bendary, ``Scheduling and communication schemes for decentralized federated learning,'' in \emph{2022 32nd International Conference on Computer Theory and Applications (ICCTA)}, 2022, pp. 122--128, doi: 10.1109/ICCTA58027.2022.10206255.

\bibitem{Nguyen21}
D.~C. Nguyen, M.~Ding, Q.-V. Pham, P.~N. Pathirana, L.~B. Le, A.~Seneviratne, J.~Li, D.~Niyato, and H.~V. Poor, ``Federated learning meets blockchain in edge computing: Opportunities and challenges,'' \emph{IEEE Internet of Things Journal}, vol.~8, no.~16, pp. 12\,806--12\,825, Aug. 2021, doi: 10.1109/jiot.2021.3072611.

\bibitem{Shi23}
Y.~Shi, L.~Shen, K.~Wei, Y.~Sun, B.~Yuan, X.~Wang, and D.~Tao, ``Improving the model consistency of decentralized federated learning,'' Feb. 2023, doi: 10.48550/ARXIV.2302.04083.

\bibitem{Giuseppi22}
A.~Giuseppi, S.~Manfredi, and A.~Pietrabissa, ``A weighted average consensus approach for decentralized federated learning,'' \emph{Machine Intelligence Research}, vol.~19, no.~4, pp. 319--330, Jul. 2022, doi: 10.1007/s11633-022-1338-z.

\bibitem{Du23}
M.~Du, H.~Zheng, X.~Feng, Y.~Chen, and T.~Zhao, ``Decentralized federated learning with markov chain based consensus for industrial iot networks,'' \emph{IEEE Trans. on Industrial Informatics}, vol.~19, no.~4, pp. 6006--6015, April 2023.

\bibitem{Schmid20}
R.~Schmid, B.~Pfitzner, J.~Beilharz, B.~Arnrich, and A.~Polze, ``Tangle ledger for decentralized learning,'' in \emph{2020 IEEE Int. Parallel and Distributed Processing Symposium Workshops (IPDPSW)}.\hskip 1em plus 0.5em minus 0.4em\relax IEEE, May 2020, pp. 852--859, doi: 10.1109/ipdpsw50202.2020.00144.

\bibitem{Neto23}
H.~N.~C. Neto, J.~Hribar, I.~Dusparic, D.~M.~F. Mattos, and N.~C. Fernandes, ``A survey on securing federated learning: Analysis of applications, attacks, challenges, and trends,'' \emph{IEEE Access}, vol.~11, pp. 41\,928--41\,953, 2023, doi: 10.1109/ACCESS.2023.3269980.

\bibitem{Wang20}
H.~Wang, K.~Sreenivasan, S.~Rajput, H.~Vishwakarma, S.~Agarwal, J.~yong Sohn, K.~Lee, and D.~Papailiopoulos, ``Attack of the tails: Yes, you really can backdoor federated learning,'' Jul. 2020, doi: 10.48550/ARXIV.2007.05084.

\bibitem{Shi22}
J.~Shi, W.~Wan, S.~Hu, J.~Lu, and L.~Y. Zhang, ``Challenges and approaches for mitigating byzantine attacks in federated learning,'' Dec. 2022, doi: 10.48550/ARXIV.2112.14468.

\bibitem{Rajput20}
S.~Rajput, H.~Wang, Z.~Charles, and D.~Papailiopoulos, ``Detox: A redundancy-based framework for faster and more robust gradient aggregation,'' Jul. 2020, doi: 10.48550/ARXIV.1907.12205.

\bibitem{Park21}
\BIBentryALTinterwordspacing
J.~Park, D.-J. Han, M.~Choi, and J.~Moon, ``Sageflow: Robust federated learning against both stragglers and adversaries,'' in \emph{Advances in Neural Information Processing Systems}, M.~Ranzato, A.~Beygelzimer, Y.~Dauphin, P.~Liang, and J.~W. Vaughan, Eds., vol.~34.\hskip 1em plus 0.5em minus 0.4em\relax Curran Associates, Inc., 2021, pp. 840--851. [Online]. Available: \url{https://proceedings.neurips.cc/paper\_files/paper/2021/file/076a8133735eb5d7552dc195b125a454-Paper.pdf}
\BIBentrySTDinterwordspacing

\bibitem{Yin18}
D.~Yin, Y.~Chen, K.~Ramchandran, and P.~Bartlett, ``Byzantine-robust distributed learning: Towards optimal statistical rates,'' 2018, doi: 10.48550/arXiv.1803.01498.

\bibitem{Sattler20}
F.~Sattler, K.-R. Muller, T.~Wiegand, and W.~Samek, ``On the byzantine robustness of clustered federated learning,'' in \emph{ICASSP 2020 - 2020 IEEE Int. Conf. on Acoustics, Speech and Signal Processing (ICASSP)}.\hskip 1em plus 0.5em minus 0.4em\relax IEEE, May 2020, doi: 10.1109/icassp40776.2020.9054676.

\bibitem{Li24}
\BIBentryALTinterwordspacing
S.~Li, E.~C.-H. Ngai, and T.~Voigt, ``An experimental study of byzantine-robust aggregation schemes in federated learning,'' \emph{IEEE Transactions on Big Data}, p. 1–13, 2024, doi: 10.1109/tbdata.2023.3237397. [Online]. Available: \url{http://dx.doi.org/10.1109/TBDATA.2023.3237397}
\BIBentrySTDinterwordspacing

\bibitem{Xu23}
Z.~Xu, Y.~Zhang, G.~Andrew, C.~A. Choquette-Choo, P.~Kairouz, H.~B. McMahan, J.~Rosenstock, and Y.~Zhang, ``Federated learning of gboard language models with differential privacy,'' May 2023, doi: 10.48550/ARXIV.2305.18465.

\bibitem{Cao22}
X.~Cao, M.~Fang, J.~Liu, and N.~Z. Gong, ``Fltrust: Byzantine-robust federated learning via trust bootstrapping,'' Dec. 2022, doi: 10.48550/ARXIV.2012.13995.

\bibitem{Guo21}
\BIBentryALTinterwordspacing
H.~Guo, H.~Wang, T.~Song, Y.~Hua, Z.~Lv, X.~Jin, Z.~Xue, R.~Ma, and H.~Guan, ``Siren: Byzantine-robust federated learning via proactive alarming,'' in \emph{Proc. of the ACM Symp. on Cloud Computing}, ser. SoCC '21.\hskip 1em plus 0.5em minus 0.4em\relax New York, NY, USA: Association for Computing Machinery, 2021, p. 47–60, doi: 10.1145/3472883.3486990. [Online]. Available: \url{https://doi.org/10.1145/3472883.3486990}
\BIBentrySTDinterwordspacing

\bibitem{Montgomery11}
\BIBentryALTinterwordspacing
D.~Montgomery, C.~Jennings, and M.~Kulahci, \emph{Introduction to Time Series Analysis and Forecasting}, ser. Wiley Series in Probability and Statistics.\hskip 1em plus 0.5em minus 0.4em\relax Wiley, 2011. [Online]. Available: \url{https://books.google.es/books?id=-qaFi0oOPAYC}
\BIBentrySTDinterwordspacing

\bibitem{LeCun10}
\BIBentryALTinterwordspacing
C.~C. Yann~LeCun and C.~J. Burges, ``The mnist database of handwritten digits,'' 2010. [Online]. Available: \url{http://yann.lecun.com/exdb/mnist/}
\BIBentrySTDinterwordspacing

\bibitem{LeCun98}
Y.~LeCun, L.~Bottou, Y.~Bengio, and P.~Haffner, ``Gradient-based learning applied to document recognition,'' \emph{Proceedings of the IEEE}, vol.~86, no.~11, pp. 2278--2324, 1998, doi: 10.1109/5.726791.

\end{thebibliography}
\end{document}